\documentclass[preprint,12pt,
]{elsarticle}
\usepackage{amssymb}
\usepackage{tikz}
\usepackage{subfig}
\usepackage{multirow}
\usepackage{lineno}
\usepackage{hyperref}
\usepackage{tabularx}
\usepackage{smartdiagram}
\usepackage{rotating}
\usepackage{longtable}
\usepackage{lscape}
\usepackage{float}
\usepackage[a4paper, total={6in, 9in}]{geometry}
\modulolinenumbers[5]
\journal{Computers in Biology and Medicine}
\newcolumntype{H}{>{\setbox0=\hbox\bgroup}c<{\egroup}@{}}
\begin{document}
	
\begin{frontmatter}
\title{Vision-based Human Fall Detection Systems using Deep Learning: A Review}
\author[add1]{Ekram Alam}
\ead{ealam4u@gmail.com}
\author[add2]{Abu Sufian}
\ead{sufian.csa@gmail.com}
\author[add3]{Paramartha Dutta}
\author[add4]{Marco Leo}
\address[add1]{Department of Computer Science, Gour Mahavidyalaya,West Bengal, India}
\address[add2]{ Department of Computer Science, University of Gour Banga, India}
\address[add3]{Department of Computer and System Sciences, Visva-Bharati University,  India }
\address[add4]{National Research Council of Italy, Institute of Applied Sciences and Intelligent Systems, 73100 Lecce, Italy.}

\begin{abstract}
Human fall is one of the very critical health issues, especially for elders and disabled people living alone. The number of elder populations is increasing steadily worldwide.   Therefore, human fall detection is becoming an effective technique for assistive living for those people. For assistive living, deep learning and computer vision have been used largely. In this review article,  we discuss deep learning (DL)-based state-of-the-arts non-intrusive (vision-based) fall detection techniques.   We also present a survey on fall detection benchmark datasets. For a clear understanding, we briefly discuss different metrics which are used to evaluate the performance of the fall detection systems.  This article also gives a future direction on vision-based human fall detection techniques.
\end{abstract}


\end{frontmatter}

\section{Introduction}
\label{sec:Intro}
\begin{figure} [h]
	\scriptsize
	\centering
	\vspace{-5pt}
	\begin{tikzpicture}[rotate=0, level 1/.style={sibling distance=70mm,level distance=15mm}, level 2/.style={sibling distance=15mm,level distance=20mm}, level 3/.style={sibling distance=11mm,level distance=40mm},
		every node/.style = {shape=rectangle, rounded corners,
			draw, align=center,
			top color=white, bottom color=blue!20, auto}]]
		
		\node {Types of Fall}
		child { 
			node {According to \\Bendary et al. \cite{el2013fall}} 
			child { node {forward }}
			child { node {backward}}
			child { node {sideway}}
		} 
		child { 
			node {According to \\Putra et al. \cite{putra2018event} }
			child { node [left=-4mm]{forward }}
			child { node [left=-5mm]{backward}}
			child { node  [left=-3mm, text width=.8 cm] {right-side}}
			child { node [left=-2mm, text width=.8 cm] {left-side}}
			child { node [left=-4mm, text width=1.3 cm]{blinded 
					forward}}
			child { node [text width=1.3 cm] {blinded backward}}
		};

	\end{tikzpicture}
	\vspace{-6pt}
	\caption{Classification of Falls by \cite{el2013fall, putra2018event}.}
	\vspace{-10pt}
	\label{F_FallTypes} 
\end{figure}
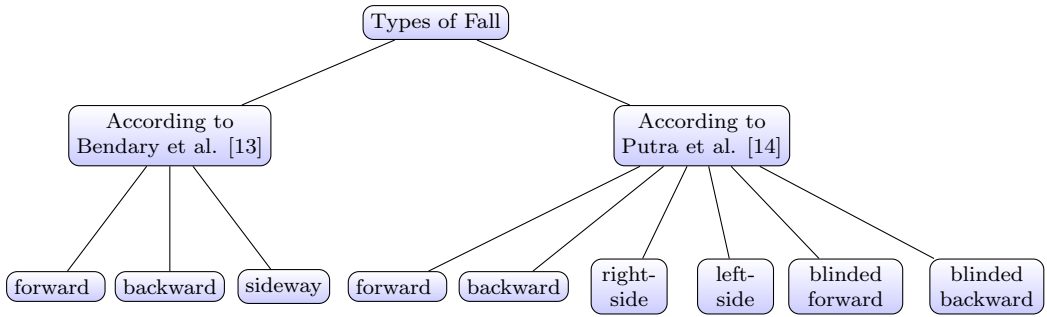

According to a report of the United Nations (UN) \cite{WPA2020H},  the elderly population of age 65 years or above was 727 million globally in 2020, which is expected to reach 1.5 billion (more than two times) by 2050. The percentage of the elderly population of this age group was 9.3\% in 2020 which will be 16.0\% in 2050. Due to a better standard of life and improvement in healthcare, the average life expectancy is increasing. If this trend continues, Very soon there will be more elderly populations than adults. So, the caring of the elders may become a big problem due to the scarcity of caretakers. These days we are very much dependent on technologies, and for the elderly population, this is not an exception. One of the main health problems in the elderly population is fall due to weakness and other reasons. Falls are one of the most common reasons for hospitalization for elderly people \cite{rougier2006monocular}. Not only prevention but detection of falls as early as possible is very crucial for the health of the concerned person.  A slight delay in detecting a fall and providing medical help can be fatal.   The chance of dying a person within 6 months after the fall is 50\% if the person is on the floor for more than one hour after the fall \cite{wild1981dangerous}, \cite{marquis2005gathering}. So, detection, prediction, and raising alarm to take action as early as possible is very important. Automatic collection and reporting of fall incidents can be used to analyze the cause of falls and can help to prevent falls. Nowadays telehealth is becoming very effective for frequent outbreaks of pandemics and epidemics \cite{garfan2021telehealth, sufian2020survey}. Fall detection can be done using wearable sensors; like gyroscopes and accelerometers; and vision sensors like RGB cameras, infrared cameras, depth cameras, and 3D-based methods using camera arrays. Wearable devices capture abnormal values like velocities and angles from the sensors and raise an alarm to alert the user and third party. One of the main problems in wearable devices is the requirement of frequent charging. For older adults, it may be difficult to continuously monitor the battery charging status and charge it frequently. Wearable sensors are also not so comfortable and have many side effects.  Due to frequent charging, uncomfortable and other side effects of the sensors this option is not so suitable for elderly people. So, non-intrusive vision-based sensors can be a good choice \cite{sufian2021h}.  Vision-based fall detection is a better alternative that provides a low-cost solution for the fall detection problem  \cite{espinosa2019vision}. Modern artificial intelligence, specifically DL \cite{alam2021leveraging} is very effective for this kind of task \cite{sultana2018advancements, jena2021artificial}. Also, due to the increased use of IoT \cite{sufian2021deep} solutions and the more uses of cameras in airports, bus stands, railway stations, roads, streets, and homes, vision-based methods for fall detection is a good choice for the future as well. Though there are many other approaches for fall detection, DL based approach is gaining momentum. As compared to other methods, there is no need of handcrafted feature extraction in DL. Automatic feature extraction is possible in DL-based methods. DL is also popular due to its generalization. A model trained on a dataset can be used for a different problem using transfer learning. The performances of DL-based techniques are also very high as compared to other methods. DL can also be used in low computing edge devices using transfer learning and few-shot learning. Fall can be of different types and duration. According to Bendary et al.  \cite{el2013fall}, fall can be of three types namely forward, backward, and sideway. Putra  et al. \cite{putra2018event} classifies fall detection into forward (ff), backward (fb), right-side (fr), left-side (fl), blinded-forward (bff), and blinded-backward (fb). Figure \ref{F_FallTypes} shows this classification. 

The rest of the paper is organized as follows. Section \ref{S_RelatedWorks} mentions some related works. Section \ref{S_EvaMetrics} discusses metrics to evaluate the performance of the different fall detection techniques. A brief description of the publicly available human fall dataset (HFDS) is provided in section \ref{S_Datasets}. Section \ref{S_Review} reviews the vision-based fall detection techniques using deep learning. Finally section \ref{S_Conclusion} conclude with some future direction.

\section{Related Works}
\label{S_RelatedWorks}

Gutierrez et al. \cite{gutierrez2021comprehensive} published a recent review work on vision-based fall detection methods. This paper reviewed all types of vision-based fall detection algorithms. Our paper specifically discusses only deep learning-based non-intrusive (vision) fall detection methods in detail. Also, we have classified our review based on different types of deep learning models. Wang et al. \cite{wang2020elderly} and Xefteris et al. \cite{xefteris2021performance}  provided a  multi-model review paper where the signals from different sensors (wearable sensors, vision sensors. ambient sensors) were fused. Our paper only reviews vision sensors. Rastogi et al. \cite{rastogi2021systematic} presented a brief machine learning-based fall detection review paper which considers all types of sensors (vision, wearable, ambient). Our paper describes only deep learning (a subset of machine learning) based fall detection techniques considering only vision sensors. In this paper, we have reviewed vision-based non-intrusive human fall detection (HFD) techniques using deep learning.

\section{Evaluation Metrics}
\label{S_EvaMetrics}
A fall detection problem can be considered a binary problem. In a binary problem, we have two outcomes with respect to each input, True (T) and False (F). For fall detection problem T means occurring a fall and F means no fall. Many works were done to detect the falls. To compare and evaluate the performances of the experiments done by different researchers,  many metrics have been designed. Some of the important metrics used by most of the researchers are given below \cite{zhu2010sensitivity}.

\begingroup
\vspace{-5pt}
\begin{tabbing}
	
	$\bullet$ Sensitivity \hspace{4em} \= $\bullet$ Specificity \hspace{5em} \= $\bullet$ Accuracy   \\
	$\bullet$ F Score \>  $\bullet$ Precision \> $\bullet$ Geometric Mean \\
	
\end{tabbing}
\vspace{-10pt}
\endgroup
All of the above-mentioned metrics can be defined on the terms of TP (True Positive), TN (True Negative), FP (False Positive), and FN (False Negative). TP means detecting falls correctly. FP denotes detecting activities of daily living (ADL) as a fall. FP means wrongly detecting a normal activity as a fall. Similarly,  FN means detecting a normal activity as a fall. \\

\textbf{Sensitivity (Recall) } is the ratio between TP and sum of TP and FN as shown in  equation \ref{E_Sens}. It is the proportion of true positive falls that are correctly identified by a fall detection system. Sensitivity signifies that how correctly an experiment can detect a actual fall as fall.  
\begin{equation} \label{E_Sens}
	Sensitivity = \frac{TP}{(TP+FN)} 
\end{equation}

Similarly, \textbf{Specificity} is the ratio between TN falls and sum of TN falls and FP falls as shown in equation \ref{E_Specif}. It signifies that how correctly an experiment can detect a normal activity as normal activity.
\begin{equation} \label{E_Specif}
	Specificity = \frac{TN}{(TN+FP)} 
\end{equation}

Sensitivity and Specificity do not give the full information about the accuracy of an experiment. To know how accurate an experiment is, both sensitivity and specificity are important. High sensitivity and low specificity and high specificity and low sensitivity are not so accurate. For, good accuracy both sensitivity and specificity should be high. \textbf{Accuracy} combines both sensitivity and specificity as shown in equation \ref{E_Acc}. 

\begin{equation} \label{E_Acc}
	Accuracy = \frac{(TP + TN)}{(TP + TN + FP + FN)} 
\end{equation}

Similarly, \textbf{Precision} can be defined mathematically as shown in equation \ref{E_Pre}. It is the ratio of TP falls to all positive falls (TP and FP).
\begin{equation} \label{E_Pre}
	Precision = \frac{TP}{(TP+FP)} 
\end{equation}

F Score (FS), also known as the F1 score is the harmonic mean of precision and sensitivity. Mathematically, it can be defined as shown in equation \ref{E_FS}. FS combines the features of precision and recall.

\begin{equation} \label{E_FS}
	FS  = \frac{2}{(\frac{1}{recall} \times \frac{1}{precision} )} 
	= 2 \times \frac{precision \times recall}{ precision + recall}
\end{equation}

Some researchers have also used \textbf{Geometric Mean (GM)} \cite{li2017fall,  charfi2013optimized} of sensitivity and specificity for the analysis of their experiments. The GM can be calculated using the following equation.

\begin{equation} \label{E_GM}
	GM = \sqrt{(SE * SP)}
\end{equation}

\section{Benchmark Public Human Fall Datasets}
\label{S_Datasets}
DL-based techniques are generally very data-hungry. It requires lots of data for training, testing, and validation. Benchmark HFDS can be utilized for this purpose. Benchmark HFDSs are also required for the evaluation and comparison of different HFD techniques. The publicly available benchmark HFDSs \cite{auvinet2010multiple, charfi2013optimized, kwolek2014human, ma2014depth, sucerquia2017sisfall, martinez2019up, maldonado2019fallen} have been created by performing fall and ADL activities by some subjects. Many researchers also used some pre-trained DL models. These models were pre-trained  on other human activity recognition (HAR) datasets like KTH \cite{schuldt2004recognizing}, UCF-101 \cite{soomro2012ucf101}, MSRDailyActivity3D \cite{wang2012mining} etc. and other large size datasets like ImageNet \cite{deng2009imagenet}, Microsoft (MS) COCO \cite{lin2014microsoft}, NTU RGB+D 120 \cite{liu2019ntu}.The Table \ref{T_Datasets} summaries the details of HFDSs. The full forms of the acronyms used in Table \ref{T_Datasets} are given in Table \ref{T_HFDSAcr}.  A brief description of the HFDSs are given below. 

\subsection{Multiple Cameras Fall Dataset (MCFD)}

Auvinet et al. \cite{auvinet2010multiple} introduced a dataset using 8 IP cameras which contains total 24 activities. Each activity was recorded using 8 different cameras from different viewpoints simultaneously. The recordings are available in AVI format.  The fall and ADL activities were performed by a single subject. Falling and other activities recorded in this dataset are given below.

\begingroup
\scriptsize
\vspace{-10pt}
\begin{tabbing}
	\hspace{2em}1. Walking \hspace{7em} \= 2. Falling \hspace{5em} \= 3. Lying on the ground \hspace{4em} \=4. Crouching  \\
	\hspace{2em}5. Moving down \> 6. Moving up \>7. Sitting \> 8. Lying on a sofa\\
	\hspace{2em}9. Moving horizontally \>10. Standing up\\
\end{tabbing}
\vspace{-18pt}
\endgroup

\subsection{Le2i Fall Detection Dataset (Le2i FDD)}
Charfi et al. \cite{charfi2013optimized} presented fall detection dataset \cite{charfi2013optimized} using a single RGB camera. Originally they named it S, now commonly known as the Le2i FDD dataset. Nine subjects performed 3 different types of fall activities (forward falls, balance loss, falls from sitting) and 6 different ADL activities (sitting, walking, standing, moving a chair, housekeeping) and captured 143 fall videos and 79 ADL videos. Videos were captured in four different environments (Home, Coffee room, Office, Lecture Room). Activities were performed in varying the various factors like light, clothes, the color of clothes, texture of clothes, shadows, reflections, camera view, etc. 

\subsection{University of Rzeszow Fall Detection (URFD) Dataset}

URFD dataset \cite{kwolek2014human} was presented by Kwolek and Kepski in 2014. This dataset contains 40 ADL sequences and 30 fall sequences. Two MS Kinect cameras were used for capturing the depth and RGB images. This dataset also contains accelerometer data.

\subsection{SDUFall}
Ma et al. \cite{ma2014depth} introduced a depth fall dataset named SDUFall \cite{ma2014depth} using an MS Kinect camera. Ten subjects (male and female) performed the following six actions.

\begingroup
\scriptsize
\vspace{-10pt}
\begin{tabbing}
	\hspace{1em}1. Falling down \hspace{3em} \= 2. Walking \hspace{5em} \= 3. Bending  \\
	\hspace{1em}4. Lying \> 5. Squatting \> 6. Sitting\\
	
\end{tabbing}
\vspace{-19pt}
\endgroup
Each subject repeated an action 30 times with variations in light, room layout, camera angle and position, carrying an object, etc. Total 1800 (6 x 30 x 10) video clips of 8-second duration were recorded using an MS Kinect camera installed at a height of 1.5 meters. The files were saved in AVI format. 

\subsection{High Quality Fall Simulation Dataset (HQFSD)}
Baldewijns et al. \cite{baldewijns2016bridging}  presented the HQFSD \cite{baldewijns2016bridging} using five web cameras (640 x 480 resolution, 12 fps) recorded in nursing home environments. Ten subjects performed 55 fall scenarios of average length 2:45 min (min length 50 s, max length 4:58 min). The total duration of fall recording by each camera is 2:25:54 hours. Fall scenarios differ in the use of walking aid (walker, wheelchair, no walking aid), fall speed (slow, fast), moving objects during the fall (blanket, walker, wheelchair, chair, none), starting pose (standing, sitting, squatting, bending over, transitions), ending positions (lying on the floor, getting back up after fall), etc. Total 17 different ADL scenarios were performed of average length 20:39 min (min length 11:38 min, max length 35:30 min). The total duration of ADL recording by each camera is 5:50:29 hours. The following 13 ADL scenarios were recorded. The videos were saved in AVI format.

\begingroup
\scriptsize
\vspace{-10pt}
\begin{tabbing}
	\hspace{1em}1. Sitting \hspace{13em} \= 2. Eating and drinking \hspace{2em} \= 3. Picking something from the floor  \\
	\hspace{1em}4. Transition \> 5. Making the bed\> 6. Putting and removing shoes\\
	\hspace{1em}7. Sleeping \> 8. Changing clothes \> 9. Coughing and sneezing  \\
	\hspace{1em}10. Reading \> 11. Walking \> 12. Getting into and out of bed \\
	\hspace{1em}13. Wheelchair to chair and vice versa 
	
\end{tabbing}
\endgroup

\subsection{Thermal Simulated Fall dataset (TSFD)}
Vadivelu et al. \cite{vadivelu2016thermal} presented a TSFD \cite{vadivelu2016thermal} using a single FLIR ONE thermal camera. The camera was mounted on an Android phone. Fall activities were done in a room environment. TSFD contains 9 ADL video segments (VS) and 35 fall VS.

\subsection{SisFall}
Sucerquia et al. \cite{sucerquia2017sisfall} presented a fall dataset named SisFall \cite{sucerquia2017sisfall}. This dataset was basically created using two accelerometers and a gyroscope. SisFall also contains video sequences \cite{euprazia2020novel}. It has 19  ADL (Walking slowly, Stumble while walking, Jogging slowly et.) video sequences and 15 (forward, lateral, vertical, etc.) fall video sequences. 

\begin{table}[h] 
	\scriptsize
	\caption{The full forms of the acronyms used in Table \ref{T_Datasets} \label{T_HFDSAcr}}
	
	\vspace{-8pt}	
	\setlength\tabcolsep{1.5pt}					
	\begin{tabularx}{\linewidth}{lXlX}  
		
	
	CN  \space  $\rightarrow$& No. of Cameras & SN \space \space $\rightarrow$ & No. of Subjects\\ 
	MP  \space$\rightarrow$& Multi-persons & Occ. $\rightarrow$ & Occlusion\\ 
	FAN $\rightarrow$&No. of Fall Activities &AN \space \space $\rightarrow$ & No. of ADL Activities \\ 
	TAN $\rightarrow$&Total No. of Activities &FDN $\rightarrow$ & No. of Fall Data Sequences\\ 
	ADN $\rightarrow$& No. of ADL Data Sequences&TD \space \space  $\rightarrow$  & Total No of Data Sequences

\end{tabularx}

\vspace{-10pt}					
\end{table}


\begin{table}[h]
\scriptsize
\caption{Benchmark Human Fall Datasets}
\vspace{-10pt}
\label{T_Datasets}	
\setlength\tabcolsep{1.5pt}
\begin{tabularx}{\linewidth}{ |p{1.5cm}|p{.8cm}|p{1.5cm}|p{1.2cm}|p{.6cm}|p{.6cm}|p{.6cm}|p{.6cm}|p{.7cm}|p{.6cm}|p{.7cm}|p{1.1cm}|p{1.1cm}|p{1.2cm}|X|} \hline
	
	\textbf{Dataset}&\textbf{Year}&\textbf{Sensor Type}&\textbf{Camera Type}    &\textbf{CN} &\textbf{SN}&\textbf{MP} &\textbf{Occ.}&\textbf{FAN }&\textbf{AN} & \textbf{TAN} &\textbf{FDN} &\textbf{ADN} & \textbf{TD} & \textbf{Link} \\ \hline

		Le2i FDD \cite{charfi2013optimized} & 2013 & Vision & RGB  & 1 &9 & No & Yes & 3 & 6 & 9& 143 VS &79 VS& 222 VS & \href{ http://le2i.cnrs.fr/Fall-detection-Dataset}{Link} \\ \hline
	
	URFD \cite{kwolek2014human}&2014&Vision & Depth, RGB     &2 &1 &No & No &- &-&-&30 VS& 40 VS&70 VS& \href{http://fenix.univ.rzeszow.pl/~mkepski/ds/uf.html}{Link} \\ \hline
	
	SDUFall \cite{ma2014depth}& 2014 &Vision & Depth & 1 & 10&No & No & 1& 5& 6&300 VS & 1500 VS&1800 VS & \href{http://www.sucro.org/homepage/wanghaibo/SDUFall.html}{Link}\\ \hline
	
	HQFSD \cite{baldewijns2016bridging} & 2016 & Vision & Web Camera & 5 &10 & Yes & Yes & 24&13& 37 & 120 VS&65 VS&185 VS& \href{http://www.kuleuven.be/advise/datasets}{Link} \\ \hline
	
	TSFD \cite{vadivelu2016thermal}& 2016 & Vision & Thermal & 1 & - & - &-& -&-&-&35 VS &9 VS&44& \href{ https://drive.google.com/open?id=0ByBHFkIRDnx6S2M2WllKaVg5eGc}{Link} \\ \hline
	
	SisFall \cite{sucerquia2017sisfall}& 2017& Wearable, Vision & RGB  & 1 &15 & No & No & 15& 19& 34&15 &19 &34 & \href{http://sistemic.udea.edu.co/en/research/projects/english-falls}{Link}\\ \hline
	
	UP-Fall Detection \cite{martinez2019up}&2019 &Wearable, Ambient, \& Vision &RGB (PNG) & 2 &17&No & No & 5& 6& 11 &510 VS& 612 VS&1122 VS& \href{https://sites.google.com/up.edu.mx/har-up/}{Link}\\ \hline

	FPDS \cite{maldonado2019fallen} & 2019 & Vision & RGB & 1 & & Yes & No & -& - &-& 1072 Images& 1262 Images& 2062 Images & \href{ http:
		//agamenon.tsc.uah.es/Investigacion/gram/papers/fall_detection/FPDS_dataset.zip.}{Link}\\ \hline

\end{tabularx}
\vspace{-15pt}
\end{table}

\subsection{UP-Fall Detection Dataset }

Martinez et al. \cite{martinez2019up} presented a multi-modal UP-Fall Detection dataset using wearable, ambient, and vision sensors. Seventeen subjects of different ages (18-24), genders (9 males and 8 females), and weights (mean 66.8 kg) performed 11 different activities (6 ADL, 5 Fall). Each activity was repeated three times. The ADL activities performed are given below. 
\begingroup
\scriptsize
\vspace{-5pt}
\begin{tabbing}
\hspace{1em} \= 1. Walking \hspace{3em} \= 2. Standing \hspace{3em} \= 3. Jumping  \\
\>	4. Sitting \> 5. Laying\> 6. Picking up an object\\

\end{tabbing}
\vspace{-10pt}
\endgroup

The Fall activities performed are given below.

\begingroup
\scriptsize
\vspace{-9pt}
\begin{tabbing}
\hspace{1em} \= 1. Falling backward \hspace{3em} \= 2. Falling forward using hands \hspace{4em} \= 3. Falling forward using knees \\
\> 4. Falling sideward 
\> 5. Falling sitting in an empty chair

\end{tabbing}
\vspace{-14pt}
\endgroup
\subsection{Fallen Person Dataset (FPDS) }

Maldona et al. \cite{maldonado2019fallen} introduced a fall dataset named FPDS \cite{maldonado2019fallen} using a single camera fitted in a robot at a height of 76 cm. FPDS contains 1072 falls and 1262 ADL (Standing, Walking, Lying, Sitting, etc) manually labeled images. More than one subject can be in a single image. The height of subjects was from 1.2 m to 1.8 m. The activities were done in 8 different environments with varying conditions of light, shadows, reflections, etc.

\section{Review on Recent State of the art}
\label{S_Review}
\begin{figure}[h]
	\vspace{-7pt}
	\centering
	
	\includegraphics[scale=0.55]{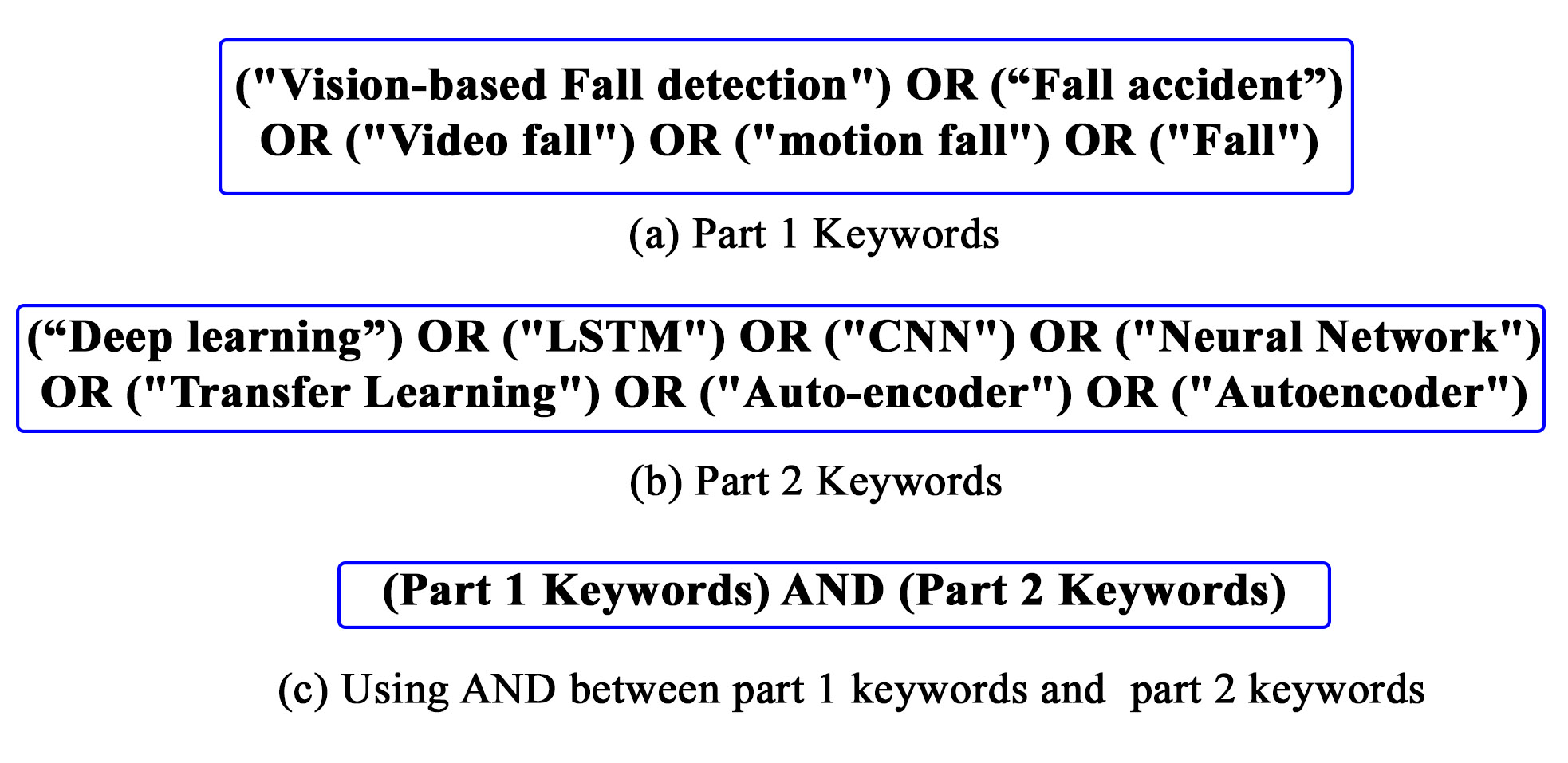}
	\vspace{-10pt}
	\caption{Generation of papers for review}
	
	\label{F_Keywords}
	\vspace{-3pt}
\end{figure}

Though research on vision-based fall detection using Neural Network has started from the year 2012 \cite{humenberger2012embedded}, the first vision-based fall detection work using deep learning is done by Feng et al. \cite{feng2014deep} in the year 2014. In this paper, we have reviewed the DL-based non-intrusive (vision) HFD methods since 2014. We have selected the different papers from Web of Science, Scopus, Google Scholar using different permutations of searching keywords. We have divided the keywords into two parts. In first part we used the keywords related to falls(like ``video fall", ``fall accident", ``motion fall" etc). In second part we used the deep learning related keyword (like ``deep learning", ``CNN", ``auto-encoder", ``LSTM" etc). We separated the keywords of 1$^{st}$ part and 2$^{nd}$ part using `OR'. We inserted `AND' between the 1st part and 2nd part keywords to generate different permutations. We filtered the generated papers manually to fit into the scope of this work. This is shown in Figure \ref{F_Keywords}.

We have classified the HFD techniques based on the DL model as given below.



\begingroup
\footnotesize
\vspace{-5pt}
\begin{tabbing}
	\hspace{2em} \= $\bullet$ Convolutional Neural Network (CNN) \hspace{2em} \= $\bullet$ Long Short Term Memory (LSTM)  \\
	\> $\bullet$ Auto-encoder  \> $\bullet$ Multi Layer Perceptron (MLP) \\
	\> $\bullet$ Hybrid
	
\end{tabbing}
\vspace{-3pt}
\endgroup

\begin{figure}[h]
	\vspace{-3pt}
	\centering
	
	\includegraphics[scale=0.55]{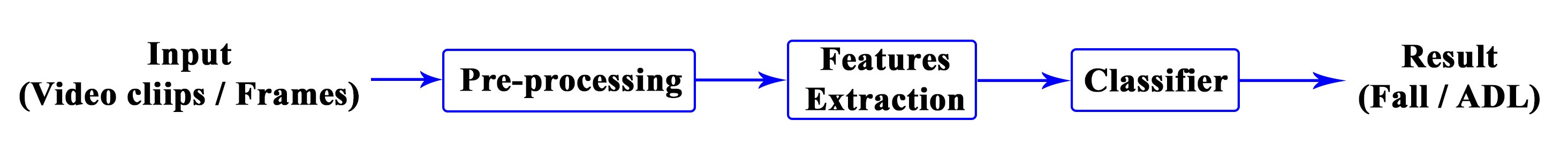}
	\vspace{-15pt}
	\caption{Basic steps in HFD}
	
	\label{F_BasicSteps}
	\vspace{-5pt}
\end{figure}

Three tables (Basic Information, Evaluation Metrics, Optimization Details) for each type of DL model have been created. The basic steps for HFD is shown in Figure \ref{F_BasicSteps}.

\subsection{CNN based techniques}
\label{SSS_FallDetectionCNN}
CNN is the one of most used DL models in computer vision and HFD is not the exception. Different sub-types of CNN are 3D CNN, TDCNN (time delay convolutional neural network), GCN (graph convolutional network), etc. The details about the experiments reviewed in this section are summarized in Table \ref{T_CNNFallDetection}, \ref{T_CNNEvaluationMetrics} and \ref{T_CNNOptimization Details}.

From the year 2016, vision-based fall detection using deep learning especially CNN gained momentum. This year, a work was reported by Doulamis \cite{doulamis2016vision} using time delay neural network (TDNN) \cite{waibel1989phoneme}.  At first, the human shape was extracted from the background, and then object tracking was applied. Here an adaptable deep learning approach was used for human detection and classification purpose. Using object tracking, an object trajectory was generated over each frame of the video. The generated object trajectory was geometrically analyzed to compute different 3D measurements to detect a human fall. Finally, A sequence of geometric features from the detected human object was fed to a TDNN to detect the human falls. A single monocular camera was used.

\begin{figure}[htb]
	\vspace{-3pt}
	\centering
	\includegraphics[scale=0.38]{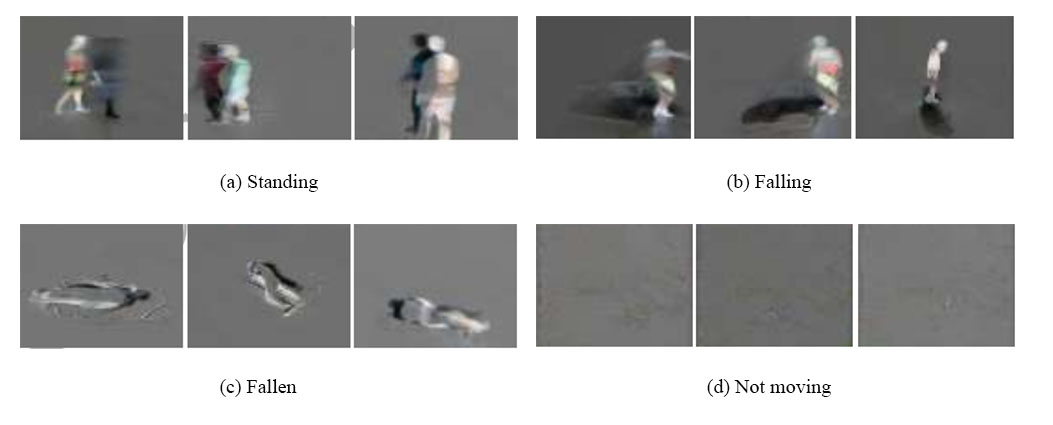}
	\vspace{-10pt}
	\caption{Some examples of dynamic images of each phase used by Fan et al. from \cite{fan2017deep}}
	\label{F_DynamicImg}
	\vspace{-3pt}
\end{figure}

Fan et al. (2017) \cite{fan2017deep} introduced fall detection method using dynamic images \cite{bilen2016dynamic}. This method not only detects the falls but also tells the duration of the fall. A fall is detected if four phases namely standing, falling, fallen, and not moving are detected in sequence. Here, the untrimmed video was decomposed into sequences of video clips. These video clips were then converted to multiple dynamic images. Dynamic images are capable of storing both the appearance and temporal information of the clips. Some examples of dynamic images have been shown in Figure \ref{F_DynamicImg}. 
  These dynamic images were fed into a deep CNN to score and classify an event as falling or not falling. To calculate the temporal extent of the fall, the difference scoring method (DSM) was used. Three publicly available datasets namely, MCFD, HQFSD, Le2i FDD, and one dataset (YouTube fall dataset (YTFD)) created by the authors were used to evaluate the experiments. YTFD dataset contained 430 falling incidents and 176 normal activities. Data augmentation was also used to increase the dataset size. To augment the data, first, they flipped each dynamic image horizontally, then cropped the image from the four corners and center to the dimension 224 $\times$ 224. A pre-trained VGG-16 CNN \cite{simonyan2014very} on ImageNet dataset \cite{deng2009imagenet} was utilized. The last fully-connected layer was replaced with a new fully-connected layer which contained four neurons. Each dataset was divided into four different non-overlapping parts. Two parts were used for training, one for testing and one for validation purposes. The initial training rate was 0.001 and it decreased to 0.1 after every 100 iterations and stopped at 300 iterations. The value of the momentum (Momt.) and weight decay (WD) were 0.9 and 0.0005 respectively. The sensitivity and specificity of the experiment used in four datasets are shown in Table \ref{T_Fan2017result}.
\begin{table}[htb] 
	\scriptsize
	\vspace{-3pt}
	\caption{Result of the experiment by Fan et al. \cite{fan2017deep} using four different data set.}
	\label{T_Fan2017result}
	\vspace{-8pt}
	\begin{tabularx}{\linewidth}{|p{3.5cm}|p{1.5cm}|p{1.5cm}|p{2.5cm}|X|} \hline
		
		\textbf{Dataset} &\textbf{Sensitivity} &\textbf{Specificity} &\textbf{Avg. Sensitivity} &\textbf{Avg. Specificity} \\ \hline
		
		Le2i FDD\cite{charfi2013optimized} &98.43 &100 &\multirow{4}{*}{83.36} &\multirow{4}{*}{83.65}\\ \cline{1-3}
		MCFD\cite{auvinet2010multiple} &97.1 &97.9 & &\\ \cline{1-3}
		HQFSD \cite{baldewijns2016bridging} &74.2 &68.6 & &\\ \cline{1-3}
		YouTube fall dataset \cite{fan2017deep} &63.7 &68.1 & &\\ \hline
		
	\end{tabularx}
	\vspace{-3pt}
\end{table}
The time to train the network was one hour. Testing time was 0.5s for converting a video clip to a dynamic image, and 0.01s for obtaining scores. 

\begin{figure}[htb]
	\vspace{-3pt}
	\centering
	\includegraphics[scale=0.3]{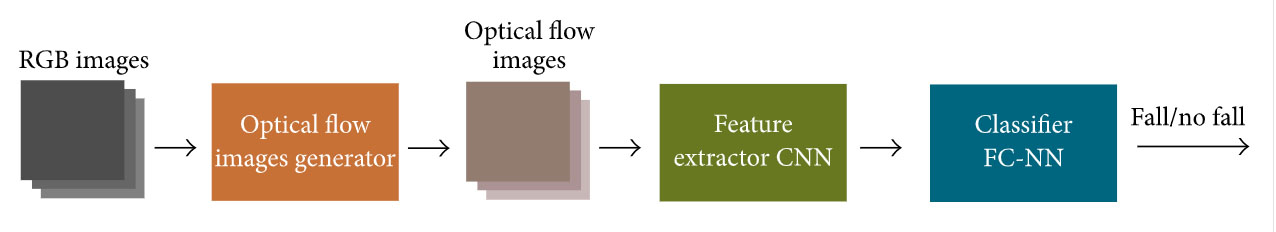}
	\vspace{-10pt}
	\caption{System architecture used by Marcos et al. \cite{nunez2017vision}}
	\label{F_marcos}
	\vspace{-3pt}
\end{figure}

Marcos et al. (2017) \cite{nunez2017vision} and Hsieh et al. (2017) \cite{hsieh2017development} utilized optical flow in their works. Marcos et al. used optical flow images \cite{horn1981determining} \cite{beauchemin1995computation} as input to the CNN network. These images only represent the motion of the consecutive frames and don't consider other appearance-related information like brightness, color, contrast, etc. Hence, using optical flow images, a generic CNN model can be built to detect the fall. A generic model can work in different scenarios. Extracted features from the generic CNN were used as input to the fully connected neural network (FC-NN) which classifies it as ``fall" or ``no fall". The steps used are shown in Figure  \ref{F_marcos}.
Here also a modified VGG-16 model was used. The input layer of VGG-16 was replaced to accept the stack of optical images as input. The size of the stack used here was 20. 
To compensate the inadequate availability of the datasets for fall detection, transfer learning \cite{pan2009survey, torrey2010transfer} was used here. Similar to Fan et al. \cite{fan2017deep}  a pretrained VGG-16 network on ImageNet \cite{deng2009imagenet} dataset was used to get a low level generic features. After that, the first layer of the network was modified to get optical flow as input. Modified network was retrained using UCF101 dataset \cite{soomro2012ucf101}. In the final steps, CNN layers up to the first fully connected layer were frozen and the last two FC layers were fine-tuned using the dropout (DO) value of 0.8 and 0.9 to get a ``fall" or ``not fall". URFD , MCFD , and the Le2i FDD	were exploited. Each dataset was split into 80:20 ratios for training and testing respectively. Adam optimization (Optim.) was used in this work. This work was implemented using the Keras framework and is available at GitHub \footnote{https://github.com/AdrianNunez/FallDetection-with-CNNs-and-Optical-Flow}. The learning rate (LR), mini batch size (MBS), activation function (AF), and other details are shown in Table \ref{T_ResultMarcos}. 
\begin{table}[h] 
	\vspace{-3pt}
	\scriptsize
	\caption{The result obtained by Marcos et al. \cite{nunez2017vision}}
	\label{T_ResultMarcos}
	\vspace{-8pt}
	\begin{tabularx}{\linewidth}{|X|p{1cm}|XH|p{2cm}|X|X|} \hline
		
		\textbf{Dataset}	& \textbf{LR}	& \textbf{MBS} &\textbf{wo} &\textbf{AF}  & \textbf{Sensitivity} & \textbf{Specificity}  \\ \hline
		
		URFD \cite{kwolek2014human} &$10 ^{-5}$& 64 & 1 & ReLU & 100 \% & 94.86 \% \\ \hline
		MCFD  \cite{auvinet2010multiple}& $10^{-3}$& Full & 1 & ReLU & 98.07 \% & 96.20 \% \\ \hline
		FDD \cite{charfi2013optimized} & $10^{-4}$ &1024& 2& ELU & 93.47\%& 97.23 \% \\ \hline
		
	\end{tabularx}
	\vspace{-3pt}
\end{table}
Hsieh et al. \cite{hsieh2017development} proposed optical flow feedback based CNN technique for fall detection. Here, the rule-based filters were used to filter out the inputs to the CNN. Falls were recognized by sequencing the frames of different actions.

Li et al. (2017) \cite{li2017fall} proposed a fall detection system where CNN was applied to each frame of the video to extract human postures. As pre-processing of the data, the average image was computed over all training and test images, and then all the training, as well as test images, are subtracted by this average image to get a uniform brightness image. After this, contrast normalization for each image was done. The architecture used, is very similar to AlexNet \cite{krizhevsky2012imagenet}. The fully connected layer FC7 and output layer fc8 were replaced by a new fully-connected layer fc7 and an output layer fc8 with 1024 and 2 neurons respectively.  ``\textbf{MatConvNet}'' \cite{vedaldi2015matconvnet}, a toolbox written using Matlab, was used for the training purpose and \textbf{Gaussian distribution} was utilized for weight parameters initialization. Minibatch stochastic gradient descent was used with batch size 85. The initial LR used was 0.05 and was multiplied by 0.1 after 20 epochs (Epo.). 40 epochs were used for training.  The publicly available dataset  ``URFD" was utilized. This dataset contains 30 fall sequences and 40 ADL sequences. The authors used all 30 fall sequences and only 28 ADL sequences out of 40. 12 ADL sequences that are recorded in the dark environment were not used. An accuracy of 99.98 was reported

Solback et al. (2017) \cite{solbach2017vision} proposed a method to detect falls using depth images captured by Stereolab’s ZED stereo cameras (left and right).  2D Human pose was estimated using CNN. MS COCO dataset was used to describe human body pose. 3D pose was generated using camera matrix, depth image, and key point information from 2D human pose estimation. Depth image was also used for 3D ground plane detection. Ground plane detection and 3D pose calculation were used for reasoning steps, which finally detects whether there is a fall or not. The true positive rate reported is 93.3 \%. This proposed system was implemented in Robot Operating System (ROS) \cite{quigley2009ros} and available publicly \footnote { https://github.com/TsotsosLab/fallen-person-detector}.

Iuga et al. (2018) \cite{iuga2018fall} proposed a fall detection system using unmanned aerial vehicle (UAV) \cite{fahlstrom2012introduction, zeggada2017deep} in an indoor setup. They used ``Parrot AR.Drone 2.0'' quadrotor \cite{hernandez2013identification} UAV which was built by a company named Parrot. YOLOv2 (You Only Look Once version 2) \cite{redmon2017yolo9000} object detection model was used to detect a human from the images captured by the AR. Drone 2.0. A pre-trained CNN with MS-COCO dataset was applied to detect a human in a standing position. Same YOLOv2 was used for fall detection with some fine-tuning in the last layer using a self created dataset (SCDS). The dataset contains 500 manually labeled images. These images were extracted from the frames of two videos. A single person wearing a fixed set of clothes was used as a  subject for these recordings. For testing, 619 images were used with two subjects both wearing different clothes than the training images. All computation was done offline using a computer on the data received by the UAV wirelessly.

\begin{figure}[htb]
	\centering
	\includegraphics[scale=0.47]{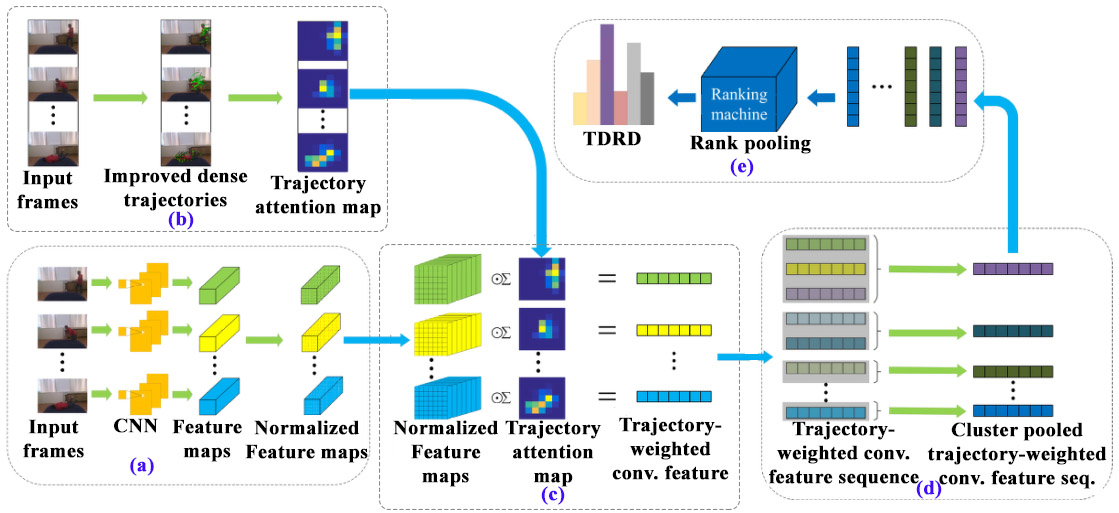}
	\vspace{-10pt}
	\caption{The process of getting TDRD as proposed  by Zhang et al.(2018). (a) Feature maps extraction and normalization. (b) Making trajectory attention maps. (c) Getting trajectory-weighted convolution feature from trajectory attention maps and normalized feature maps. (d) Cluster pooling on (c). (e) Finally getting TDRD from rank pooling on (d).}
	\label{F_Zhang18}
\end{figure}

Zhang et al. (2018) \cite{zhang2018fall} proposed a \textbf{TDRD} (Trajectory-weighted Deep-convolutional Rank-pooling Descriptor) based fall detection system. The process to get the TDRD is shown in Figure \ref{F_Zhang18}. First, CNN feature maps were extracted from the input RGB video using slightly modified VGG-16 network \cite{simonyan2014very}. Extracted feature maps were normalized using the spatio-temporal normalization method \cite{wang2015action}. Then, improved dense trajectories were calculated from the input RGB video, which was used to get trajectories attention maps. Trajectories attention maps can locate the human regions. Next, for each frame, trajectory-weighted convolution features were calculated from the extracted normalized feature maps and trajectory attention maps. This trajectory-weighted convolution feature maps sequence of all frames can give information about human action dynamics. To reduce the redundancy, cluster pooling was applied on trajectory-weighted convolutional feature sequences. Finally, rank pooling was used on the cluster pooled sequence to get the TDRD. 
The performance of the experiments and its average value using the three mentioned datasets are shown in Table \ref{T_ResultZhang}.

\begin{table}[h] 
	\scriptsize
	\caption{Results of the experiment as reported by Zhang et al.  \cite{zhang2018fall}. \label{T_ResultZhang}}
	
	\vspace{-8pt}						
	\begin{tabularx}{\linewidth}{|X|X|X|XH|}  \hline
		
		\textbf{Dataset}&\textbf{Sensitivity} &\textbf{Specificity}  &\textbf{Accuracy}& \textbf{Training Time}\\ \hline
		
		SDUFall \cite{ma2014depth}& 93.01 & 99.50 & 96.04& 65 minutes\\ \hline
		URFD \cite{kwolek2014human}& 100 &95.00 &-& 30 Minutes\\ \hline
		MCFD&90.21 &92.59 & - & 35 Minutes\\ \hline
		\textbf{	Average}&\textbf{94.41} &\textbf{95.70} &\textbf{96.04}& 43.33 Minutes \\ \hline
		
	\end{tabularx}
	
	\vspace{-3pt}					
\end{table}

\begin{figure}[htb]
	\vspace{-3pt}
	\centering
	\includegraphics[scale=0.55]{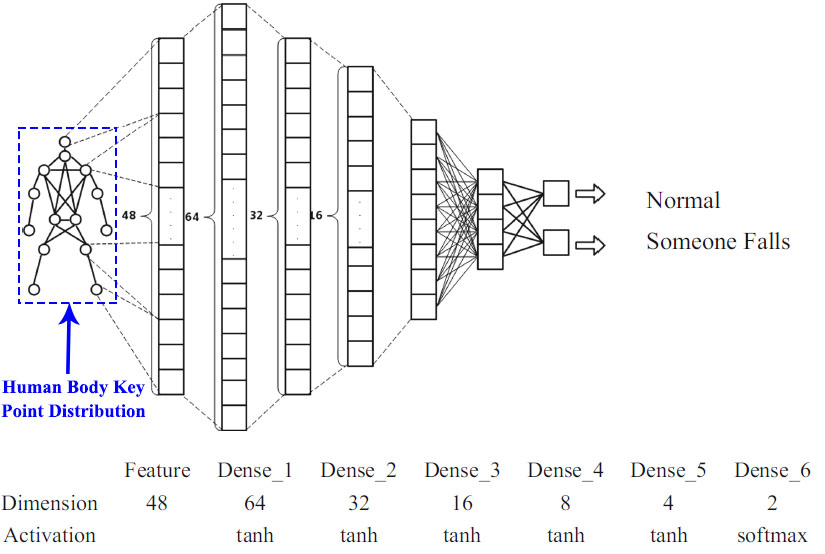}
	\vspace{-10pt}
	\caption{Depth neural network structure as proposed  by Shen et al. from \cite{shen2018fall}}
	\label{F_Shen18_Struct}
	\vspace{-3pt}
\end{figure}

Another fall detection system was presented by Shen et al. (2018) \cite{shen2018fall}. This system has three components namely (i) two cameras, (ii) a cloud server, and (iii) a smartphone. 
Cameras were used for taking the videos of the subject and converting it to high-frequency images. Raspberry Pi 3 Model B board was used for the cameras. Cloud server was used for processing the data and smartphone for receiving the alert about the falls. \textbf{Deepcut} neural network model \cite{pishchulin2016deepcut} was used to identify the 14 key points of the human body. The detected key points data map was fed into the deep neural network for the detection of falls. Figure \ref{F_Shen18_Struct} shows the structure of the depth neural network with human body key point distribution. 
The authors used 44 pre-recorded videos, out of which 42 were used for training and 2 for testing. Two cameras with various angles, scenes, and poses were used. The average accuracy reported is 98.05\%.


Kong et al. (2019) \cite{kong2019robust} showed how the height of the camera may affect the performance of the fall detection system. They used an enhanced tracking and denoising Alex-Net (ETDA-Net) which is basically an AlexNet with some pre-processing added to improve the tracking performance and to reduce the noise in the image. They created a dataset using a depth camera for their work. Images were captured from five different heights. The images captured from different heights were used for training the model. At the time of testing, ETDA-Net detected the camera height and used the model which was trained on the same height for better performance. Kaid et al. (2019) \cite{el2019reduce} used CNN architecture to reduce the false positives cases of fall detection using the angle assistance algorithm. Here the task of CNN is to filter the fall alert as false, if the image for which the fall is detected contained a person sitting in a wheelchair. Here, the authors' goal is to detect whether the fall detected image contains a person sitting in a wheelchair. If the answer is yes, then no fall alert will be generated, and if no then a fall alert will be sent. The advantage of this study is that it reduces the false-positive cases. The disadvantage is that it's not a fall detection system, it only reduces the false positives.

\begin{figure}[htb]
\vspace{-3pt}
\centering
\includegraphics[scale=0.54]{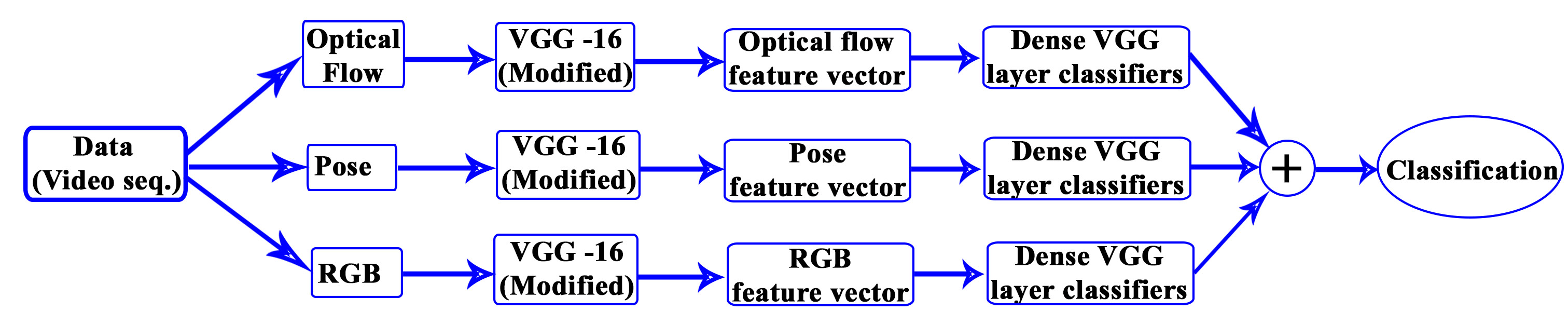}
\vspace{-10pt}
\caption{Multi-stream fall detection as proposed by Cameiro et al.}
\label{F_Cameiro}
\vspace{-3pt}
\end{figure}

\begin{figure}[h]
\vspace{-3pt}
\centering
\includegraphics[scale=0.6]{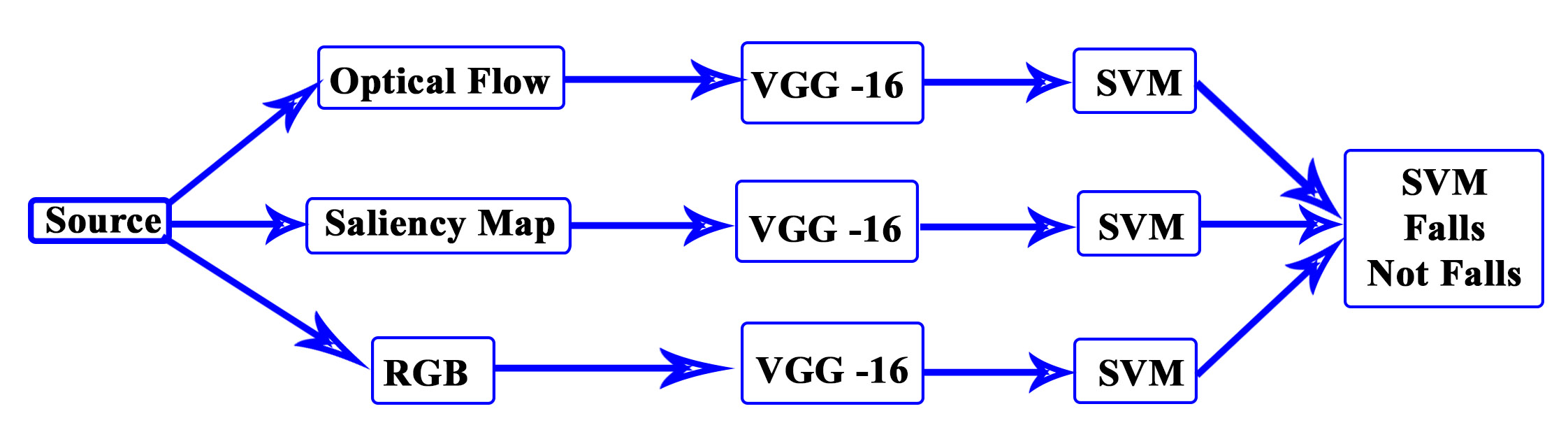}
\vspace{-10pt}
\caption{The overview of the system as proposed by Leite et al. \cite{leite2019fall}}
\label{F_Leite}
\vspace{-3pt}
\end{figure}

Cameiro et al. (2019) \cite{cameiro2019multi} and Leite et al. (2019) \cite{leite2019fall} proposed a  multi-stream fall detection method. Cameiro et al.  \cite{cameiro2019multi} exploited handcrafted high-level features. The  flow diagram of the proposed system is shown in Figure \ref{F_Cameiro}.
Three independent CNNs (Modified VGG-16) with three different inputs (optical flow, RGB, and pose) from video sequences were used. Finally, the outputs of these three CNNs were ensembled to detect a fall. ImageNet dataset was used to train each individual CNN (VGG-16). The UCF101 dataset was used for optical flow generation and training the optical flow stream.  URFD and Le2i FDD datasets were utilized for detecting a fall in video sequences. 80\% of data were used for training and 20\% for testing. The overview of the system proposed by Leite et al. \cite{leite2019fall} is shown in Figure \ref{F_Leite}. They used optical flow, saliency map, and RGB as input for the three independent VGG-16 networks. Dual SVM (Support Vector Machine) was used for the classification and for weighting each stream. ImageNet and UCF101 were used for pre-training and fine-tuning respectively. For the experiments, URFD and Le2i FDD were utilized.

\begin{figure}[h]
\vspace{-3pt}
\centering
\includegraphics[scale=0.7]{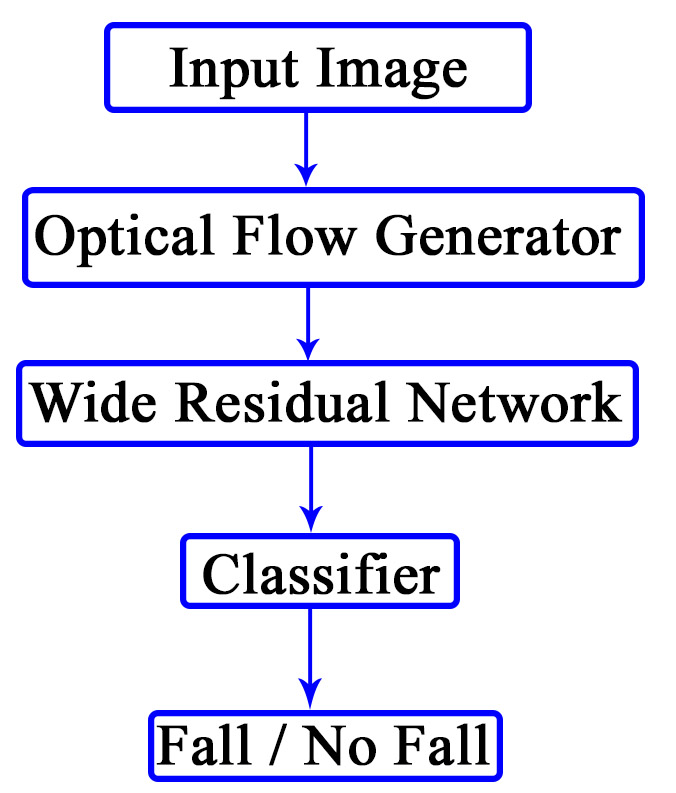}
\vspace{-8pt}
\caption{The steps of the method as proposed by Cai et al. \cite{cai2019novel}}
\label{F_CaiNovel}
\vspace{0pt}
\end{figure}

\begin{figure}[h]
\vspace{-3pt}
\centering
\includegraphics[scale=0.47]{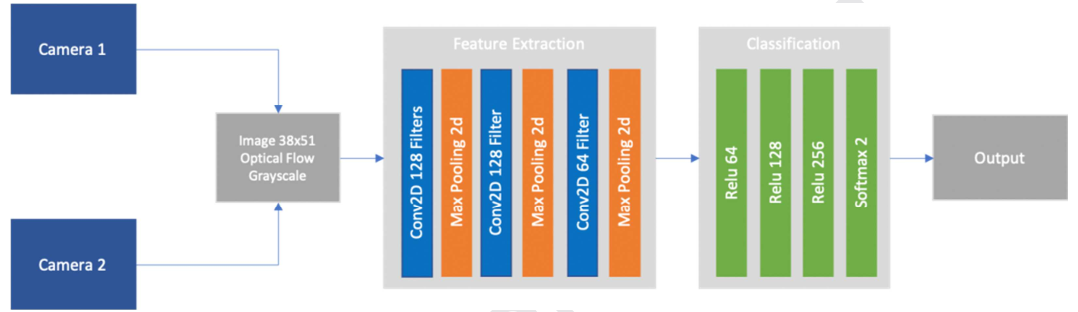}
\vspace{-10pt}
\caption{CNN model as proposed by Espinosa et al. from \cite{espinosa2019vision}}
\label{F_Espinosa}
\vspace{-1 pt}
\end{figure}

Similar to Marcos et al. \cite{nunez2017vision}, Hsieh et al. \cite{hsieh2017development}, and Cameiro et al. \cite{cameiro2019multi}; Brieva et al. (2019) \cite{brieva2019intelligent}, Cai et al. (2019) \cite{cai2019novel} and Espinosa et al. (2019) \cite{espinosa2019vision} also exploited  optical flow for fall detection. Brieva et al. \cite{brieva2019intelligent}  created a fall detection dataset using a single subject. The length of each video recording was 10 seconds. The subject performed five types of falls namely (i) falling forward using hands, (ii) falling forward using knees, (iii) falling backwards, (iv) falling sitting in empty chair, (v) falling sideward. A CNN having a single layer with 25 filters, one Relu, one FC layer, one soft function, and one classification layer was used.   The horizontal ($I_u$) and vertical components ($I_v$) of the calculated optical flow were combined as $I_{uv}$=[$I_u, I_v$] which was used as input for the CNN. The images were resized to 28 X 56 size (pixels). The total 2817 images of the dataset were divided into two groups with 1972 (70\%) images and 845 images (30\%) for training and testing respectively. The value of regularization coefficient used was 1 $\times$ $10^{-4}$. Cai et al. (2019) \cite{cai2019novel} used optical flow as input to the wide residual network. The overview of the proposed method by Cai et al. is shown in Figure \ref{F_CaiNovel}. 
Espinosa et al. (2019) \cite{espinosa2019vision} exploited a 2D CNN model as shown in Figure \ref{F_Espinosa}. Images from two cameras (lateral and front view) from the UP-fall dataset were used. Data was divided into 1-second time windows with 0.5-second overlap. Images were resized to 38x51 pixels grayscale images. 67\% (42,000 images)  of the data was used for training and 33\% (21,000 images) for the testing. The optical flow was used as a pre-processing of the data. The experiments were done on a CNN architecture designed by the authors and on VGG-16. 
Experiments were executed in Python 3.7.3 using Keras 4 and Skleran 3 frameworks.  In training, Adam optimizer was used with 50 epochs. The results of the experiments are shown in Table \ref{T_ResultEspinosa}.
\begin{table}[h!]
\vspace{-3pt}
\scriptsize
\caption{Results of the experiment as reported by Espinosa et al.  \cite{espinosa2019vision}}
\label{T_ResultEspinosa}
\vspace{-9pt}
\begin{tabularx}{\linewidth}{|l|l|p{1.9cm}|p{1.9cm}|X|} \hline
	\textbf{Camera}&\textbf{Model}&\textbf{Sensitivity} &\textbf{Specificity}  &\textbf{Accuracy}\\ \hline
	
	Cam 1(Lateral view)& Proposed CNN &97.72 &81.58 &95.24\\ \hline 
	Cam 2(Front view) & Proposed CNN &97.57 &79.67 &94.78 \\  \hline
	Cam1 \& Cam 2&Proposed CNN& 97.95 & 83.08 & 95.64\\ \hline
	Cam1 \& Cam 2&VGG-16& 100 &0 &84.44\\ \hline

\end{tabularx}

\vspace{-3pt}	
\end{table}

\begin{figure}[htb]
\vspace{-3pt}
\centering
\includegraphics[scale=0.6]{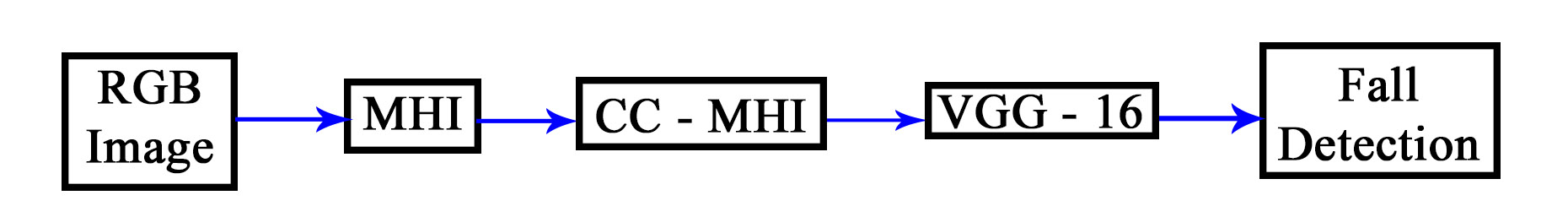}
\vspace{-10pt}
\caption{The overview of the system as proposed by Cai et al. \cite{cai2019fall}}
\label{F_Cai}
\vspace{-3pt}
\end{figure}

Cai et al. (2019) \cite{cai2019fall} proposed a fall detection system using colorization coded motion history image (CC - MHI) on VGG-16. The steps used in this method are shown in Figure \ref{F_Cai}. Firstly, MHI \cite{bobick2001recognition} was calculated using the RGB image followed by color coding of MHI. This color-coded MHI was fed as input to VGG-16. Finally, the output was generated as fall or not fall. 1000 dimensional fully connected layer of VGG-16 was replaced with 2 dimensional fully connected layers. 

\begin{figure}[h]
\vspace{-3pt}
\centering
\includegraphics[scale=0.6]{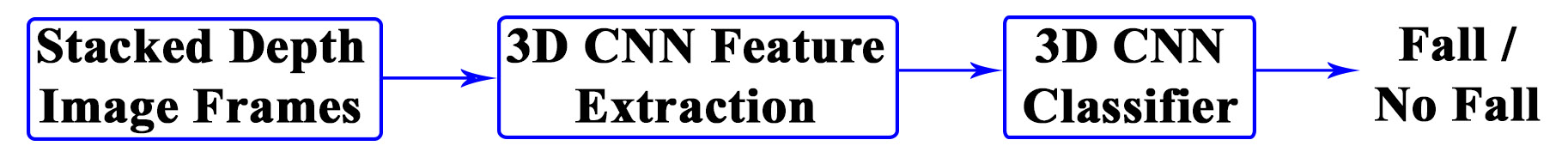}
\vspace{-8pt}
\caption{The overview of the system as proposed by Kasturi et al. \cite{kasturi2019human}}
\label{F_Kasturi}
\vspace{-1pt}
\end{figure}

\begin{figure}[htb]
\vspace{-3pt}
\centering
\includegraphics[scale=0.65]{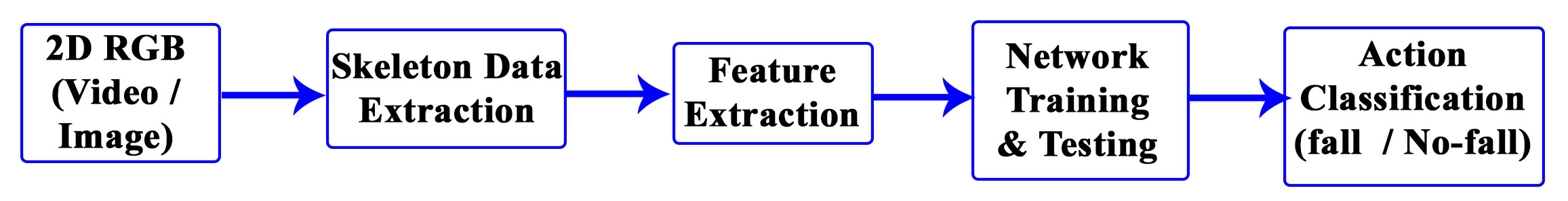}
\vspace{-10pt}
\caption{The overview of the system as proposed by Zheng and Zhou \cite{zheng2019fall}}
\label{F_Zheng19}
\vspace{-1pt}
\end{figure}

Kasturi et al. (2019) \cite{kasturi2019human} proposed a fall detection system using 3D-CNN \cite{tran2015learning}. Multi frame stacked image cube from Kinect camera was used as input to the 3D CNN. The overview of the system is shown in Figure \ref{F_Kasturi}.  
Dataset was divided into 80\% and 20\% for training and testing respectively.
Wu et al. (2019) \cite{wu2019skeleton} proposed a human skeleton sequence-based fall detection system. The sequence of motion was encoded into an RGB image.  They added their own dataset into the MSRDaily Activity3D dataset \cite{wang2012mining} to create a new dataset for their experiments. Kinect V2 was used for the data collection. All images were resized to 60 $ \times $ 60. A lightweight CNN with 4 convolutional layers, 4 max-pooling layers, and 2 FC layers, was used to classify the encoded RGB image. LeakyReLU was used as an activation function. 1200 images of fall and non-fall and 300 images of fall and non-fall were used for training and testing respectively.
Zheng and Zhou (2019) \cite{zheng2019fall} proposed a fall detection system using pose estimation and GCN. The overview of the system is shown in Figure \ref{F_Zheng19}. OpenPose \cite{cao2017realtime} was used for pose estimation. Six categories (fall, jump up, sit down, stand up, stand down, and sit down) of data from the NTU RGB+D dataset \cite{liu2019ntu} and  URFD datasets were utilized.

A very similar work to Espinosa et al. (2019) \cite{espinosa2019vision} was proposed by Espinosa et al. (2020) \cite{espinosa2020application} based on optical flow using two cameras. This work also used UP-Fall detection dataset. Carlier et al. (2020) \cite{carlier2020fall} also proposed an optical flow-based fall detector for a Nursing Home environment using modified VGG-16. Yao (2020) et al. \cite{yao2020novel} proposed geometric features based fall detection system using a shallow CNN. At first, the head was segmented from the rest of the body and two different ellipses were generated for the head and torso respectively which were used to extract the long and short axis ratio, vertical velocity, and the orientation angle features. The authors used their own dataset which contains 102 videos, captured using multiple monocular cameras from different heights and angles. 74 videos were used for training and 28 videos for testing. Different height makes this system more robust.  
Menacho et al.(2020) \cite{menacho2020fall} introduced an optical flow-based fall detection model using CNN with a reduced number of parameters that can be implemented in low computing devices like Mobile robots.

Chen et al. (2020) \cite{chen2020edge} proposed an edge computing-based fall detection technique and tried to analyze how the height and weight of the person can affect the performance. The basic parameters like height, weight, etc. were calculated using the edge node and sent to the cloud node for automatically selecting the best model. If any miss was detected, the missed image was also sent to the cloud for retraining the model.  A subpart of the thermal dataset created by Kong el at. \cite{kong2019robust}  was used. LeNet, AlexNet, and GoogleNet were exploited for the experiment. GoogleNet gave the best result.
Ijina (2020) \cite{ijjina2020human} proposed an HFD technique using temporal template representation of depth videos. Background subtraction was used as pre-processing. A pre-trained AlexNet was exploited. Hader (2020) et al. \cite{hader2020automatic} proposed a technique for fall detection using  region-based faster R-CNN \cite{ren2015faster}. Two pre-trained architecture, AlexNet and VGG-16, and three custom architecture were exploited. VGG-16 gave the best results. 

\begin{figure}[h]
\vspace{-3pt}
\centering
\includegraphics[scale=0.6] {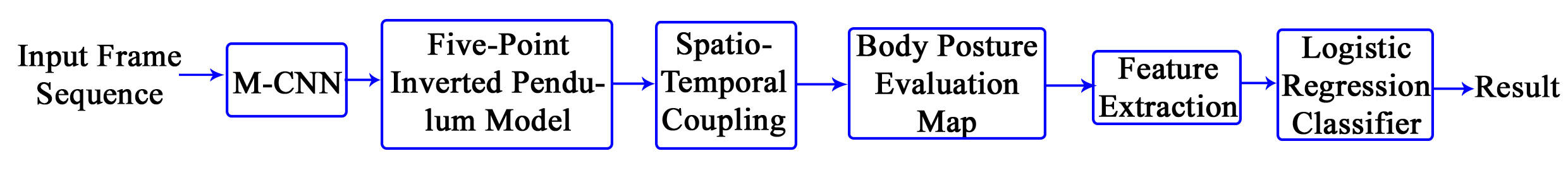}
\vspace{-13pt}
\caption{The overview of the system as proposed by Zhang et al. \cite{zhang2020human}}
\label{F_Zhang20}
\vspace{-3pt}
\end{figure}

Dichwalkar (2020) et al. \cite{dichwalkar2020activity} proposed a single shot detector (SSD) - MobileNet \cite{howard2017mobilenets}  model for fall detection. The frames of the video were resized to 300 x 300 and converted into blobs which were used as input to the SSD. For every detected object as a person, the difference of x coordinates and y coordinates were calculated. If the difference of x coordinates was greater than y coordinates then the object (person)  either fell down or is in a sleeping position. The authors used the COCO dataset for training and the test videos from  \cite{robinovitch2013video} for testing. Similar to Dichwalkar et al. \cite{dichwalkar2020activity}, Asif et al. (2020) \cite{asif2020sshfd} proposed a single shot human fall detection (SSHFD) system which also considers some missing information in pose data using occluded joins resilience (OJR). At first, a bounding box of human was generated from a single RGB image which was fed as input to a stacked hourglass (SH) Network to generate a 2D pose. Using a neural network 2d pose was converted to a 3d pose. Finally, 2d pose and 3d pose were fed to another neural network for fall detection. MS COCO  dataset was used for the training of the SH network. Authors created synthetic data was used for training the SSHFD, and for testing MCFD and the Le2i FDD were utilized. The initial LR used was 0.01 which was divided by 10 after 50\% and 75\% of the total number of epochs. Lezzar et al. (2020) \cite{lezzar2020camera} proposed a fall detection method considering more than one person in the frames and occlusions. To extract the persons (bounding boxes) YOLOv3 was used. The aspect ratio of the bounding box (AR), normalized bounding box width (NBW), and normalized bottom bounding box (NBB) were used as features for fall detection. CNN was exploited for features extraction and human detection; and SVM for recognition of posture. The FPDS, and a self-created datasets were used. Zhang et al. (2020) \cite{zhang2020human} proposed a fall detection model based on a ``five-point inverted pendulum model" using a multi-stage convolutional neural network (M-CNN). Authors created their own dataset named ``Postures of Fall (PoF)". PoF contained single or multi-person RGB images captured using a single RGB camera. The overview of the proposed model by Zhang et al. is shown in Figure \ref{F_Zhang20}.					
Zhong et al. (2020) \cite{zhong2020multi} proposed a multi-occupancy fall detection method using a thermal sensor. A pre-trained CNN was exploited. The images were divided into groups of 80\% and 20\% for the training and testing respectively. To collect the data a thermal sensor was installed on the ceiling of a room. Three persons (age: 25 to 35, one woman, two men, height: 1.68, 1.72, 1.83 meters) performed different actions for data collection. Each captured thermal image was converted to 28 x 28 size. Both single and multi-occupancy data were recorded. For single occupancy, 186 data was labeled as fallen and 159 as non-fallen. For the multi-occupancy data, 528 images were labeled as fallen and 431 as non-fallen. To increase the dataset size data augmentation was used. 

In another work, Asif et al. (2020) \cite{asif2020privacy} proposed a human pose estimation and segmentation (HPES) based fall detection technique using multiple CNN structures. They used a single camera to capture RGB images. They used synthetic data to learn human proposals using body joint locations and segmentation information to detect falls. After training on the synthetic data they also used public  MCFD and Le2i FDD to test their experiments. They represented an image to a skeleton-like structure. In this way, they removed all personal information like the face, etc of the subject and preserved the privacy. They also proposed a CNN model FallNet which uses the skeleton and segmentation-based representations and learns high-level features automatically for fall detection. The initial LR used was 0.01 which was divided by 10 after 50\% and 75\% of the total number of epochs.

\begin{figure}[h]
\vspace{-3pt}
\centering
\includegraphics[scale=0.55
]{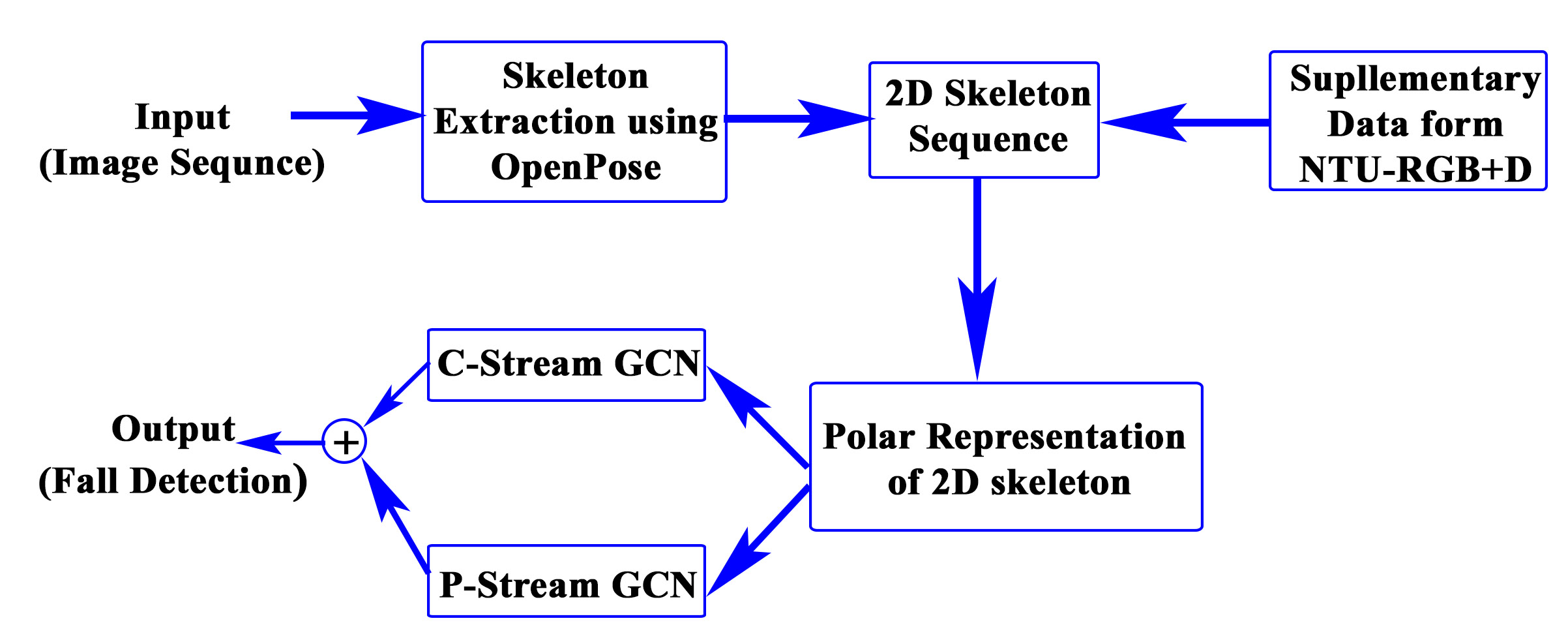}
\vspace{-10pt}
\caption{The overview of the system as proposed by Liu et al. \cite{liutwo}}
\label{F_Liu20}
\vspace{-3pt}
\end{figure}

Euprazia and Thyagharajan (2020) \cite{euprazia2020novel} proposed a method to recognize different human actions including falls. At first, action pattern image (API) was created from the input video which was fed to pre-trained series CNN (SCNN) to recognize the fall. To get the API at first edges of the frames were extracted using canny edge detection. The detected edges were combined into a single frame after proper enhancement. The LR used was 0.001 with a drop factor of 0.1 after every 8 epochs.  Liu et al. (2020) \cite{liutwo} proposed a fall detection technique based on a 2D skeleton using two-streamed GCN. The overview of the proposed system is shown in Figure \ref{F_Liu20}. At first, 2D skeleton sequence was extracted using OpenPose. Some extra data was added by converting 3D skeleton to 2D skeleton from NTU-RGB+D dataset. Then, the skeleton sequence was converted to polar form. After that, two GCN were used in parallel one for polar and another one for Cartesian. Finally, the score of these two streams' GCN was combined to get the fall result. The authors created an indoor specific action (ISA) dataset.

Chen et al. (2020) \cite{chen2020fall} and Kareem et al. (2020)  \cite{kareem2020using} presented a fall detection method using OpenPose. Chen et al. \cite{chen2020fall} used ``speed of descent at the center of hip joint'', ``the human body centerline angle'', and ``width-to-height ratio" of the bounding box to detect the falls. 
Kareem et al. \cite{kareem2020using} utilized modified optimized ResNet-50 residual network.

Abdo et al. (2020) \cite{abdo2020fall} proposed a fall detection method based on shape and motion features using RetinaNet \cite{lin2017focal} and MobileNet. RetinaNet was used to detect and track humans in video frames. Extracted handcrafted features like aspect ratio, head tracking, motion history images, head orientation were fed into MobileNet for final fall detection.

\begingroup
\vspace{-3pt}
\scriptsize
\setlength\tabcolsep{1.5pt}
\begin{longtable}{|p{3cm}|p{1.1cm}|p{3.4cm}|p{4.3cm}|p{2.6cm}|}  

\caption{Fall Detection using CNN : Basic Details\label{T_CNNFallDetection}}
\vspace{-8pt}
\endfirsthead 
\hline
\textbf{Reference}&\textbf{Sensor} &\textbf{Dataset}  &\textbf{Technique}&\textbf{Framework} \\ \hline
\endhead
\hline
\textbf{Reference}&\textbf{Sensor} &\textbf{Dataset}  &\textbf{Technique}&\textbf{Framework} \\ \hline

Doulamis et al. (2016) \cite{doulamis2016vision}&RGB &SCDS&TDNN&-\\ \hline
Fan et al. (2017) \cite{fan2017deep}&RGB &\cite{charfi2013optimized},\cite{auvinet2010multiple},\cite{baldewijns2016bridging} ,\cite{fan2017deep}& Dynamic Images, VGG-16&Caffe\\ \hline

Marcos et al. \cite{nunez2017vision} &RGB&URFD, MCFD, Le2i  FDD & Optical flow & Keras\\ \hline
Li et al. \cite{li2017fall}&RGB& URFD \cite{kwolek2014human}& Postures, Avg Image &MatConvNet\\ \hline	

Solbach et al. \cite{solbach2017vision} &RGB&MS COCO \cite{lin2014microsoft}& Pose estimation, Depth Image & ROS\\ \hline

Hsieh et al. \cite{hsieh2017development}&RGB & KTH \cite{schuldt2004recognizing}& Optical flow, rule based filter& - \\ \hline
Iuga et al. (2018) \cite{iuga2018fall}& RGB &MS COCO, SCDS&YOLO v2, UAV& -\\ \hline	
Zhang et al. \cite{zhang2018fall}& RGB & SDUFall, URFD, MCFD &TDRD&C++, Python, MATLAB\\ \hline

Shen et al. \cite{shen2018fall}&RGB  &Self created  &Deep Cut NN&Keras\\ \hline	
Kong et al. (2019) \cite{kong2019robust}&Depth &\cite{kong2019robust}&ETDA-Net& - \\ \hline	

Cameiro et al. \cite{cameiro2019multi} &RGB & Imagenet, UCF101, URFD, Le2i FDD &Multi-Stream CNNs & -\\ \hline

Brieva et al. \cite{brieva2019intelligent} &RGB &SCDS &Optical flow& - \\ \hline

Cai et al. \cite{cai2019fall} & RGB &URFD \cite{kwolek2014human}  &CC-MHI& - \\ \hline

Cai et al.  \cite{cai2019novel}&RGB & URFD \cite{kwolek2014human}  &Optical flow& - \\ \hline

Kasturi et al.  \cite{kasturi2019human} &Depth &URFD \cite{kwolek2014human}  &3D CNN&Keras, Tensorflow \\ \hline

Espinosa et al.  \cite{espinosa2019vision} &RGB&UP-fall detection\cite{martinez2019up} &Optical flow, 2D CNN& Keras 4, Sklearn 3	\\ \hline

Leite et al. \cite{leite2019fall}&RGB (Video)&ImageNet, UCF101, URFD, Le2i FDD  &Multi-Stream CNNs & - \\ \hline

Wu et al. \cite{wu2019skeleton}& RGB (Kinect) &SCDS, Activity3D dataset \cite{wang2012mining}  &Skeleton sequence, light weight CNN & - \\ \hline

Zheng and Zhou \cite{zheng2019fall}&RGB &NTU RGB+D, URFD & Pose Estimation, GCN & Pytorch \\ \hline

Espinosa et al. (2020) \cite{espinosa2020application} & RGB&UP-fall detection\cite{martinez2019up} & optical flow & - \\ \hline	

Chen et al. \cite{chen2020edge} 		
& Depth &\cite{kong2019robust}&Edge Computing & -  \\ \hline

Menacho et al. \cite{menacho2020fall}& RGB &URFD \cite{kwolek2014human}&optical flow&Keras \\ \hline

Carlier et al. \cite{carlier2020fall}&RGB& MCFD, URFD, Le2i FDD & Optical flow, custom VGG-16& -  \\ \hline

Yao et al. \cite{yao2020novel}&RGB &SCDS&Head Segmentation, Shallow CNN& - \\ \hline

Ijina \cite{ijjina2020human}&Depth &SDUFall\cite{ma2014depth}&temporal template representation& - \\ \hline

Hader  et al.  \cite{hader2020automatic}& RGB&Le2i FDD &Region based, faster R-CNN&Matlab \\ \hline

Dichwalkar  et al.  \cite{dichwalkar2020activity}& RGB& COCO, Robinovitch et al. \cite{robinovitch2013video}& SSD-MobileNet&Caffe \\ \hline

Lezzar et al. \cite{lezzar2020camera}& RGB&FPDS, SCDS &CNN and SVM& - \\ \hline
Zhang et al. \cite{zhang2020human} & RGB&SCDS (PoF) &M-CNN& - \\ \hline
Asif et al. \cite{asif2020sshfd}& RGB &COCO, SCDS, MCFD, Le2i FDD &SSHFD, OJR, SH&Torch Lib \cite{paszke2017automatic} \\ \hline	
Asif et al.\cite{asif2020privacy}& RGB&COCO, SCDS, MCFD, Le2i FDD&HPES, FallNet&Torch Lib \cite{paszke2017automatic} \\ \hline	
Euprazia and Thyagharajan \cite{euprazia2020novel}& RGB&MCFD, URFD , Le2i FDD, SisFall &SCNN with transfer learning& Matlab2018a \\ \hline
Zhong et al. \cite{zhong2020multi}& Thermal&Ulster University &Multi-occupancy& - \\ 	\hline
Liu et al. \cite{liutwo}& RGB&NTU-RGB, ISA &Two stream GCN&MMSkeleton \cite{mmskeleton2019} \\ \hline	
Chen et al. \cite{chen2020fall}&RGB &Self created&OpenPose, Speed, Angle, Ratio& \\ \hline

Abdo et al. \cite{abdo2020fall}&RGB &URFD \cite{kwolek2014human}, FDD \cite{charfi2013optimized} &RetinaNet, Handcrafted features, MobileNet&Keras, Tensorflow \\ \hline
Kareem et al.  \cite{kareem2020using} &RGB &MCFD&OpenPose, ResNet-50&Google Colab \\ \hline	
Chhetri et al. (2021)\cite{chhetri2021deep}&RGB &URFD \cite{kwolek2014human}, FDD, MCFD& Dynamic Optical Flow&Python \\ \hline

Chen et al. \cite{chen2021video}&RGB &NTU RGB+D \cite{liu2019ntu}&Pose estimation& - \\ \hline 
Killian et al.  \cite{killian2021fall} &RGB&SCDS &Cluttering awareness, YoloV3&NAOqi, Python, OpenCv \\  \hline
Leite et al.  \cite{leite2021three} &RGB &URFD \cite{kwolek2014human}, FDD \cite{charfi2013optimized}&Three stream 3D CNN& SciKit, OpenCV, Keras, TensorFlow\\ \hline 
Zou et al. \cite{zou2021movement}&RGB &Le2i FDD, MCFD, LSST \cite{zou2021movement}&3D CNN&Caffe \\ \hline 
Vishnu et al. \cite{vishnu2021human} &- &Le2i FDD, URFD, Montreal &ResNet-101& - \\ \hline 

Cai et al. \cite{cai2021vision} &RGB &URFD \cite{kwolek2014human}&MCCF, dense block& - \\ \hline

Keskes and Noumier \cite{keskes2021vision}  &Skeleton data&NTU RGB-D, TST v2, Fallfree & ST-GCN, transfer learning & Pytorch  \\ \hline

Berlin et al. \cite{berlin2021vision} &RGB &URFD , Le2i FDD  &Siamese CNN&Pyhton \\ \hline
Li et al. \cite{li2021fall}&RGB &URFD, NTFD \cite{li2021fall} &Saliency maps& Matlab 2014, Caffe \\ \hline

\end{longtable}
\endgroup
%

%
%
%
%

%
%
%
\begingroup

\scriptsize

\setlength\tabcolsep{1.5pt}			
\begin{longtable}{|p{4.7cm}|p{1.6cm}|p{1.6cm}|p{1.6cm}H|p{1.4cm}HH|p{1.4cm}|p{1.5cm}|} 
\caption{Fall Detection using CNN: Evaluation Metrics}
\label{T_CNNEvaluationMetrics}
\vspace{-3pt}
\endfirsthead 
\hline

	%

\textbf{References}&\textbf{Sensitivity} &\textbf{Specificity}  &\textbf{Accuracy}&\textbf{GM}&\textbf{F Score} &\textbf{RT}&\textbf{PvPr}&\textbf{Precision}&\textbf{Notes}\\ 
\endhead \hline

\textbf{References}&\textbf{Sensitivity} &\textbf{Specificity}  &\textbf{Accuracy}&\textbf{GM}&\textbf{F Score} &\textbf{RT}&\textbf{PvPr}&\textbf{Precision}&\textbf{Notes}\\ \hline
Fan et al.\cite{fan2017deep}&83.36* & 83.65 *& 83.86&-&-&Yes&-&-&*Average\\ \hline		
Marcos et al.\cite{nunez2017vision} (URFD)&100.0 & 92.00 & 95.00&-&-&-&-&-&-\\ \hline
Marcos et al.\cite{nunez2017vision} (MCFD)&99.00 & 96.00 &-&-&-&-&-&-&- \\ \hline
Marcos et al.\cite{nunez2017vision} (FDD)&99.00 & 97.00 & 97.00 &- &- &- &- &- &- \\ \hline	
Li et al.\cite{li2017fall}&100.0 &99.98& 99.98&99.99&0.0234&yes&No&-&- \\ \hline
Solbach et al. \cite{solbach2017vision}& - & - & Above 91 &- &-  &- &- &- &-\\ \hline

Iuga et al. \cite{iuga2018fall}&86.4 &-&-&-&-&-  &- &- &-\\ \hline

Zhang et al. \cite{zhang2018fall}&94.41* &95.70*& 96.04*&G-&-&-&No&-&*Average\\ \hline
Shen et al.\cite{shen2018fall}&- &-&98.05 &-&- &-&-&-&-\\ \hline	

Cameiro et al. \cite{cameiro2019multi} (FDD)  &99.9 &98.32& 98.43&-&-&-&-&-&-\\ \hline
Cameiro et al. \cite{cameiro2019multi} (URFD)&100 &98.61& 98.77&-&-&-&&-&-\\ \hline
Cameiro et al.(Average) \cite{cameiro2019multi}&99.95 &98.46& 98.6&-&-&-&-&-&-\\ \hline

Brieva et al. \cite{brieva2019intelligent}&95.42 &- &92.78&-&95.34&RT&PvPr&95.27&-\\ \hline
Brieva et al.(using majority voting) \cite{brieva2019intelligent}&95.42 &- &96.84&GM&-&RT&PvPr&95.27&- \\ \hline

Cai et al.  \cite{cai2019fall}&- &- &92.34&GM&- &RT&PvPr&-&-\\ \hline

Cai et al.  \cite{cai2019novel}&- &- &92.6&GM&-&RT&Yes&-&-\\ \hline

Kasturi et al. \cite{kasturi2019human}&- &- &100*&GM&-&RT&PvPr&- & *maximum\\ \hline

Espinosa et al. \cite{espinosa2019vision}&97.95 &83.08 &-&-&-&-&-&-&- \\ \hline

Leite et al.  \cite{leite2019fall}(URFD)&100&98.77 &98.84&GM&-&RT&PvPr&-&-\\ \hline
Leite et al.  \cite{leite2019fall}(FDD)&99.43&98.55 &99.51&GM&-&RT&PvPr&-&-\\ \hline

Wu et al. \cite{wu2019skeleton}&93.9 &- &93.75&GM&-&RT&PvPr&-&-\\ \hline

Zheng and Zhou \cite{zheng2019fall}&97.1 &- &94.1&GM&-&RT&PvPr&94.4&-\\ \hline

Espinosa et al. \cite{espinosa2020application}&97.95&83.08&95.64&gm&-&rt&pp&96.91&-\\ \hline

Chen et al. \cite{chen2020edge}&92.884*&99* &	97.352* &gm&-&rt&pp&-& *Average\\ \hline

Menacho et al. \cite{menacho2020fall}		
&41.47 &- &88.55&GM&-&Yes&PvPr&31.043&-\\ \hline

Carlier et al. (URFD) \cite{carlier2020fall} &96.7&-&-&gm&-&rt&pp&85.3&-\\ \hline
Carlier et al. (FDD) \cite{carlier2020fall} &90.9&-&-&gm&-&rt&pp&92.8&-\\ \hline
Carlier et al. (MCFD) \cite{carlier2020fall} &71.0&-&-&gm&-&rt&pp&87.1&-\\ \hline
Carlier et al. (Avg) \cite{carlier2020fall} &86.2&-&-&gm&-&rt& pp&88.4&-\\ \hline
Yao  et al.\cite{yao2020novel}&- &- &90.5&GM&-&Yes&PvPr&-&-\\ \hline

Ijina\cite{ijjina2020human}&- &- &91.58&GM&-&&Yes&-&-\\ \hline
Hader  et al. (VGG-16) \cite{hader2020automatic}&99.4466 &- &99.5&GM&-&-&PvPr&99.526&-\\ \hline
Dichwalkar  et al. \cite{dichwalkar2020activity}&- &- &60&GM& -&- &PvPr&- &-\\ \hline
Lezzar et al. \cite{lezzar2020camera}&100 &- &&GM&-&-&PvPr&93.94 &-\\ \hline	
Zhang et al.\cite{zhang2020human}&- & - &98.7&GM&- &- &PvPr&-  &-\\ \hline	

Asif et al.(MCFD) \cite{asif2020sshfd}&84.31 &- &- &GM&84.53*&&PvPr&84.87& *F1 score\\ \hline
Asif et al.(Le2i) \cite{asif2020sshfd}&89.92 &- &- &GM&89.91*&&PvPr&90.08& *F1 score\\ \hline

Asif et al.(MCFD) \cite{asif2020privacy} &98.61* &- &-&GM&98.60* &&PvPr&98.60*&*Maximum\\ \hline
Asif et al.(Le2i)\cite{asif2020privacy}&92.44* &- &- &GM&92.44*&&PvPr&92.45* & *Maximum\\ \hline

Euprazia and Thyagharajan \cite{euprazia2020novel}&- &-  &100&GM&- &-& PvPr&-  &-\\ \hline	
Zhong et al. \cite{zhong2020multi}&- &- & $95.89^*$  &95.92 (±0.68)&-&- &PvPr&- &$^*$ (±0.50)\\ \hline
Liu et al.\cite{liutwo}&- &- &- &GM&- &&PvPr&-  &-\\ \hline

Chen et al.  \cite{chen2020fall}&98.3 &95 &97&GM&- &RT&PvPr&-  &-\\ \hline	
Abdo et al.  \cite{abdo2020fall}&97.7* &100 &98&-&- &-&&-& *Approx\\ \hline	

Kareem et al. \cite{kareem2020using}&- &- &97.46&GM&-&RT&PvPr&- &-\\ \hline	
Chhetri et al. \cite{chhetri2021deep}&- &- &91.405*&GM&- &RT&PvPr&- &*Average\\ \hline	
Chen et al. \cite{chen2021video}&- &- &99.83&GM&- &RT&PvPr&- &-\\ \hline	
Killian et al. \cite{killian2021fall}&- &- &88.88&GM&- &RT&PvPr&- &-\\ \hline	
Leite et al. \cite{leite2021three} (URFD)&0.99 &- &0.99&GM&- &RT&PvPr&- &-\\ \hline	
Leite et al. \cite{leite2021three} (FDD)&0.99 &- &0.99&GM&- &RT&PvPr&-  &-\\  \hline
Zou et al.  \cite{zou2021movement}&100.00 &97.04 &97.23&GM&-&RT&PvPr&- &-\\ \hline					
Vishnu et al. \cite{vishnu2021human} (Le2i) &93.0 &- &- &GM&86.8&RT&PvPr&81.5 &-\\ \hline
Vishnu et al. \cite{vishnu2021human} (URFD)&76.6 &- &- &GM&82.1&RT&PvPr&88.4 &- \\ \hline
Vishnu et al. \cite{vishnu2021human} (Montreal)&99.1 &94.8 &- &GM&- &RT&PvPr&- &-\\ \hline
Cai et al. \cite{cai2021vision}&95.0 &100 &96.6&GM&97.3&RT&PvPr&100 &-\\ \hline

Keskes and Noumier  \cite{keskes2021vision} (TST v2)&- &- &100&GM&-&RT&PvPr&- &-\\ \hline
Keskes and Noumier  \cite{keskes2021vision} (Fallfree)&- &- &97.33&GM&- &RT&PvPr&- &-\\ \hline
Berlin et al.  \cite{berlin2021vision} (URFD) &- &- &100&GM&-&RT&PvPr&-&-\\ \hline
Berlin et al.  \cite{berlin2021vision} (FDD) &- &- &97&GM&-&RT&PvPr&-&-\\ \hline
Li et al.  \cite{li2021fall} (URFD)&- &- &99.67&GM&-&RT&PvPr&-&-\\ \hline
Li et al.  \cite{li2021fall} (NTFD)&- &- &98.92&GM&-&RT&PvPr&-&-\\ \hline	
%

\end{longtable}
\vspace{-3pt}
\endgroup

\begingroup				

\scriptsize
\setlength\tabcolsep{1.5pt}
\begin{longtable}{ |p{3.8cm}|p{1.3 cm}|p{1cm}|p{1.3cm}|p{1cm}|p{1cm}|p{1cm}|p{1.2cm}|p{.7cm}|p{1.4cm}|}  
\caption{Fall Detection using CNN: Optimization Details}
\vspace{-8pt}
\label{T_CNNOptimization Details}	
\endfirsthead \hline
\textbf{Reference}&\textbf{Optimiz.} &\textbf{MBS}  &\textbf{LR} &\textbf{Momt.}&\textbf{WD}&\textbf{Epo.}&\textbf{Classi-fier}&\textbf{DO}&\textbf{AF}\\ \hline
\endhead \hline
\textbf{References}&\textbf{Optimz.} &\textbf{MBS}  &\textbf{Lr.Rt.} &\textbf{Momt.}&\textbf{W.D.}&\textbf{Epoch}&\textbf{Classi-fier}&\textbf{DO}&\textbf{Act.Fn.}\\ \hline

Fan et al.\cite{fan2017deep}& - & -&0.001&	0.9&0.0005&300&Softmax & - & -	\\ \hline	
Marcos et al.\cite{nunez2017vision} &Adam &64 to 1024 &0.00001, 0.001, 0.0001&- & -& 3000-6000&FC-NN& 0.9, 0.8&ReLU, ELU \\ \hline

Li et al.\cite{li2017fall}&SGD&85&0.05 &0.9&0.0001&40&- & - & - \\ \hline	
Iuga et al. \cite{iuga2018fall}&SGD&4 &0.0001 &0.99&- & - & - & - & -\\ \hline	
Shen et al.\cite{shen2018fall}&-&-&-&-&-&-& Softmax & -& tanh \\ \hline
Cameiro et al.  \cite{cameiro2019multi}&Adam&powers of 2&$10^{-3}$ to $10^{-5}$&0.99&-&500 & - & -&-\\ \hline

Brieva et al. \cite{brieva2019intelligent} &SGD &16 & 0.01&0.9&-&500& Softmax & & ReLU\\ \hline

Cai et al.  \cite{cai2019fall}&SGD  &- & 0.0005&-& -&-& Softmax & 0.5  & -  \\ \hline

Cai et al.  \cite{cai2019novel}&- &16 to 256  &$10^{-3}$ to $10^{-5}$&- & -& -& Softmax&-&ReLU\\ \hline

Kasturi et al. \cite{kasturi2019human}&SGD &8 & 0.001&-&-&100&-&0.2 & -\\ \hline
Espinosa et al. \cite{espinosa2019vision} &Adam&-&-& -&- & 50 &Softmax& -&ReLU \\ \hline 
Leite et al.  \cite{leite2019fall}&Adam &1024  & 0.0001&-&-&500&SVM & - & -\\ \hline

Wu et al. \cite{wu2019skeleton}&Adam &32 & -&-&-&-& Softmax&-&Leaky ReLU\\ \hline

Menacho et al. \cite{menacho2020fall}& RMS-prop &32 & 0.001&-&-&20&-&0.5& ReLU\\ \hline
Carlier et al. \cite{carlier2020fall} &op& 128, 256&0.001 &-&-&-&-& -& ReLU \\ \hline 
Yao  et al.\cite{yao2020novel}&SGD &- & 0.00001&-&-&500&Softmax&-&ReLU\\
 \hline

Ijina\cite{ijjina2020human}&- &- & -&-&-&-&SVM&- & -\\ \hline
Hader  et al. (VGG-16) \cite{hader2020automatic}&SGD &- & 0.001&-&-&10&Softmax&-& ReLU\\ \hline
Lezzar et al. \cite{lezzar2020camera}&-& -& - & -& -&- &SVM&-&- \\ \hline	
Asif et al.\cite{asif2020sshfd}&Adam &- & 0.01  &-&0.0005&300&-&0.5& ReLU\\ \hline 

Asif et al. \cite{asif2020privacy}&Adam &- &  0.01  &-&0.0005&150&Softmax&- & -\\ \hline
Euprazia and Thyagharajan \cite{euprazia2020novel}&SGD &45 & 0.001&0.9&-&30&Softmax&- & -\\ \hline

Zhong et al. \cite{zhong2020multi}&Adam &32 & 0.001&-&-&-&Softmax&-& ReLU\\ \hline
Liu et al.\cite{liutwo}&SGD &64 & 0.01&0.9&0.0001&9&Softmax&- & -\\ \hline

Abdo et al.  \cite{abdo2020fall}&-&32 &0.01&0.95&0.001&200& Softmax &-&ReLU  \\ \hline				
Kareem et al. \cite{kareem2020using}&RMS-Prop &- & -&-&-&10&Softmax&0.2&ReLu\\ \hline
Chhetri et al. (2021)\cite{chhetri2021deep}&- &- &-&-&-&-&Softmax&- & -\\ \hline
Chen et al. \cite{chen2021video}&Adam &- & 0.0001&-&-&20&-&0.25&ReLU\\ \hline 
Leite et al. \cite{leite2021three}(FDD)&Adam &192 & 0.00001&-&-&500&SVM&0.5 & -\\ \hline 

Zou et al. (2021) \cite{zou2021movement}&SGD &8 & &0.9&0.00005&-&Softmax&-& tanh\\ \hline 

Keskes and Noumier  \cite{keskes2021vision}&SGD &- & 0.01&-&-&100&Softmax&0.5 & -\\ \hline
Berlin et al.  \cite{berlin2021vision}&- &- & -&-& - &30& - & - & ReLU, Sigmoid\\ \hline 
Li et al.  \cite{li2021fall}& - & - & - & - & - & - & - & - &ReLU, Sigmoid\\ \hline


\end{longtable}
\endgroup	

	Chhetri et al. (2021) \cite{chhetri2021deep} proposed a fall detection system using enhanced dynamic optical flow. Three benchmark datasets namely  URFD, MCFD, and Le2i FDD were used. At first, Input video was converted into RGB images. Then, the optical flow was calculated using those converted RGB images. After that, Rank pooling was calculated that was used to calculate the dynamic flow. Using the dynamic flow, the dynamic image was created. These steps were done as pre-processing. After pre-processing stage,  feature extraction and classification were done. For this, a pre-trained VGG-16 (modified) was used. To make the fall detection more challenging, a variation of light was added to the input data. Results are shown in Table \ref{T_ResultChhetri}.

	\begin{table}[h]
		\vspace{-3pt}
		\scriptsize
		\caption{Results of the experiment as reported by Chhetri et al. (2021) \cite{chhetri2021deep}}
		\vspace{-8pt}
		\label{T_ResultChhetri}
		\centering						
		\begin{tabular}{|l|l|} \hline

			\textbf{Dataset}&\textbf{Accuracy}\\ \hline
			
			URFD \cite{kwolek2014human}& 95.11\\ \hline
			MCFD \cite{auvinet2010multiple} & 92.91 \\ \hline
			FDD \cite{charfi2013optimized}& 91.1\\ \hline
			FDD \cite{charfi2013optimized} (Dark lighting condition)&86.5\\ \hline
			\textbf{	Average}& \textbf{91.405} \\ \hline

		\end{tabular}
		
		\vspace{-3pt}	
	\end{table}

	Chen et al. (2021) \cite{chen2021video} introduced a pose-based fall detection technique. First of all, a 2D pose was extracted which was then converted to 3D pose. This 3D pose was fed to CNN for fall detection. NTU RGB+D dataset was exploited. The samples which have more than one person or some missing values were not used. Killian et al. (2021) \cite{killian2021fall} proposed fall prevention and detection technique using a bipedal humanoid NAO robot \cite{gouaillier2009mechatronic}. To prevent the fall a cluttering awareness technique was used. For this, the current image of the room of interest was compared with the standard reference image. If the cluttering level is high NAO robot speaks and asks to arrange the objects. To detect a fall YOLOv3 was used. If very little movement is detected in continuous images then there is a chance of fall. In this case, NAO robot asks some pre-defined questions to verify whether it's an actual fall or a false fall. If the person says that he/she needs helps or no response is received, then the robot sends the alerts to the caregivers using the Internet. The average accuracy in a lab environment was 94.44. The average highest and lowest accuracy in the real home environment was 88.88 and 83.33 respectively. The overall average accuracy was 88.88.

	Leite et al. (2021) \cite{leite2021three} introduced three-streamed 3D CNN network-based fall detection system. A pre-trained network on ImageNet was used for each stream. Three pre-trained models were trained again using Optical flow, Visual Rhythm, and pose estimation features independently. Finally, an SVM classifier was used for fall detection.  URFD and Le2i FDD datasets were utilized with augmentation. Dataset was divided into 65\%, 15\%, and 20\% for training, validation, and testing respectively. Zou et al. (2021) \cite{zou2021movement} presented fall detection method using 3D CNN. At first, the sequence of frames was fed as input to the 3D CNN to extract 3D features. A reshape layer converts the 3D features to the 2D features. Then, a spatial pyramid was used to generate multi-scale features which were fed to the movement tube regression layer, tube anchors generation layer, and softmax classification layer. After that, a matching and hard negative mining layer were used to find the tube anchors that match the ground truth using intersection-over-union (IOU). Finally, falls were detected using softmax and movement tube regression layer.  SSD is used to detect objects. Apart from Le2i FDD \cite{charfi2013optimized} and MCFD \cite{auvinet2010multiple}, a large-scale spatial-temporal (LSST) dataset was created. Data augmentation was used by changing the spatial, illumination, and temporal properties.
	Vishnu et al. (2021) \cite{vishnu2021human} proposed a fall motion vector based fall detection method using 3D CNN. A pre-trained ResNet-101 was used. Le2i \cite{charfi2013optimized}, URFD \cite{kwolek2014human} and Montreal \cite{charfi2012definition} datasets were  used. Berlin et al. (2021) \cite{berlin2021vision} proposed a siamese neural network \cite{koch2015siamese, he2018twofold} based fall detection technique using single-shot classification \cite{vinyals2016matching}. Siamese network has two symmetrical convolutional networks. These two networks calculate the difference of input video sequence from another ADL video sequence and generate a distance value in the range of [0,1] where 1 denotes complete similarity and 0 denotes no similarity. Two filters, traditional 2D Conv. filter, and depth-wise Conv. filter were used. Two types of input features stacked RGB features, and optical flow features were utilized.

	Cai et al. (2021) \cite{cai2021vision} introduced a multi-channel convolutional fusion (MCCF) based fall detection method using dense blocks and transition layers. At first, input frames (10) were fed to the convolutional layer. The output of the convolutional layer was given to the MCCF-Dense block. The result of the dense block was fed to the transition layer for down-sampling and reducing the data redundancy. After a few repetitions of dense blocks and transition layers, a final dense block was used. Finally, a fully connected layer was added. The output of the FC layer was fed to softmax for final fall detection. Li et al. (2021) \cite{li2021fall} introduced fused saliency-based fall detection system. This system had two parts namely the saliency maps generation part and fall detection part. Two-stream CNN model one for global stream and another one for local stream were used for saliency maps generation. Generated fused saliency maps were fed as input to a five-layer CNN for final fall detection. 
	
	Keskes and Noumier (2021) \cite{keskes2021vision} proposed a fall detection system using spatial-temporal graph convolutional networks (ST-GCN) \cite{yan2018spatial}. NTU RGB-D \cite{liu2019ntu}, TST v2 \cite{cippitelli2016tst}, Fallfree \cite{alzahrani2017fallfree} dataset were used with transfer learning technique. The ST-GCN network consisted of ST-GCN units where each unit contains a spatial GCN and temporal GCN. The ST-GCN was consists of 10 layers. The first four layers consisted of 64 output channels, the next three layers contained 128 output channels and the last three layers consisted of 256 output channels. The kernel size used was 9. The ST-GCN was pre-trained on NTU RGB-D dataset. After that, the first nine layers were frozen and retrained using TST v2 and fallfree datasets. The input data (Skeleton graph) was represented as tensor (N,3,T,25,M) where N is the batch size, 3 is the joint coordinates(x,y,z), T is the number of frames in one video clip, 25 is the total number of skeleton joints of each person and M is the number of person in the same clip.  (N,3,T,25,M) was modified to (N*M, C,T,V) and was fed to spatial GCN to extract the spatial features. The extracted spatial features were fed to temporal GCN to extract the temporal feature vectors of the same joint. Finally, softmax was used on these feature vectors to detect falls.

	
	%

\subsection{LSTM or RNN (Recurrent Neural Network) based Techniques}
\begin{figure}[h]
	\vspace{-10pt}
	\begin{center}
		\includegraphics[scale=0.5]{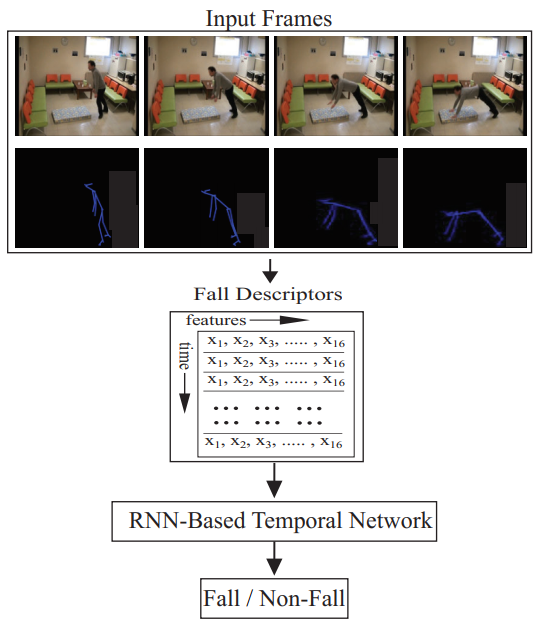}
		\vspace{-10pt}
		\caption{The overview of the system as proposed by Hasan et al. from \cite{hasan2019robust}}
		\label{F_Hasan}
	\end{center}
	\vspace{-3pt}
\end{figure}

\begin{figure}[h]
	\vspace{-3pt}
	\centering
	\includegraphics[scale=0.4]{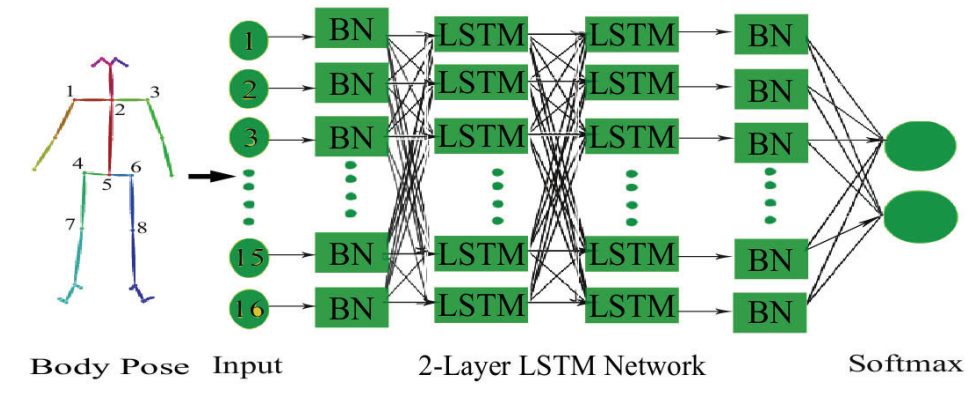}
	\vspace{-12pt}
	\caption{The overview of the architecture as proposed by Hasan et al. from \cite{hasan2019robust}}
	\label{F_Hasan2}
	\vspace{-1pt}
\end{figure}

In this section LSTM based fall detection techniques are reviewed. The details about the experiments discussed in this section are summarized in Table \ref{T_RNNFallDetection}, \ref{T_RNNEvaluationMetrics} and \ref{T_RNNOptimization Details}. Hasan et al. (2019) \cite{hasan2019robust}  proposed a  method to detect a fall from video data using RNN with 2 layers LSTM. Here, the 2D pose estimation using OpenPose \cite{cao2017realtime} algorithm was done which gave body joints. The body joints (RKnee, LKnee, RHip, MidHip, LHip, RShoulder, LShoulder, and Neck) were selected and a 24 frame pose sequence timestep was used. After that, extracted pose vectors were fed into a 2-layer LSTM which finally detected a fall. The overview of the system as proposed by Hasan et al. is shown in Figure \ref{F_Hasan}.
For every 24 frame sequence an overlap of 16 frames (66.67\%) was used. The architecture used is shown in Figure \ref{F_Hasan2}.
For the 8 joints total 16 values were required. Batch normalization(BN) was used after input and after the last LSTM. 

\begin{table}[h]
	\centering
	\vspace{-3pt}
	\scriptsize
	\caption{Fall Detection using RNN: Basic Details}
	\label{T_RNNFallDetection}
	\vspace{-10pt}
		\setlength\tabcolsep{1.5pt}
		\begin{tabular}{|p{3.7cm}H|p{3.5cm}|p{5.9cm}|p{1.7cm}|} \hline
			
			\textbf{Reference}&\textbf{Sensor} &\textbf{Dataset}  &\textbf{Technique}&\textbf{Framework} \\ \hline

			Hasan et al. (2019)  \cite{hasan2019robust} &RGB &MCFD, URFD, Le2i FDD&Pose estimation with LSTM &- \\ \hline

			Jeong et al. \cite{jeong2019human}&RGB &URFD, SDUFall  &Skeleton extraction, SHCLC&TensorFlow\\ \hline

			Feng et al. (2020) \cite{feng2020spatio}&RGB & SCDS, URFD, MCFD& Attention guided LSTM, YOLO V3, VGG-16 & -\\ \hline

			Romaissa et al. \cite{romaissa2020fall} &RGB &Le2i FDD &Body Geometry, Optical flow&- \\ \hline
			
			Taufeeque et al. (2021) \cite{taufeeque2021multi}& -&UP-Fall, SCDS&Pose estimation&PyTorch \\  \hline
			
		\end{tabular}
		\vspace{-3pt}	
	\end{table}					
	\begin{table}[h]
		\centering
		\vspace{0pt}
		\scriptsize
		\caption{Fall Detection using RNN : Evaluation metrics}
		\label{T_RNNEvaluationMetrics}
		\vspace{-7pt}						
			
			\begin{tabular}{ |l|p{1.5cm}|l|p{1.5cm}H|lHH|l|} \hline
				
				\textbf{References}&\textbf{Sensitivity} &\textbf{Specificity}  &\textbf{Accuracy}&\textbf{GM}&\textbf{F Score} &\textbf{RT}&\textbf{PvPr}&\textbf{Precision}\\ \hline

				Hasan et al. (MCFD) \cite{hasan2019robust}&98.0 &96.0 &-&GM&-&RT&PvPr &- \\ \hline
				
				Hasan et al. (Le2i FDD) \cite{hasan2019robust}&99.0 &97.0 &-&GM&-&RT&PvPr &- \\ \hline
				
				Hasan et al. (URFD) \cite{hasan2019robust}&99.0 &96.0 &-&GM&-&RT&PvPr &- \\ \hline
				
				Hasan et al. (Average)&98.67 &96.33 &-&GM&-&RT&PvPr &- \\ \hline

				Jeong et al. \cite{jeong2019human}&-&- &98.83&GM&-&RT&PvPr &- \\ \hline

				Feng et al. \cite{feng2020spatio} (SCDS) &83.5&-&-&gm&86.5&rt&pvpr&89.8\\ \hline
				
				Feng et al. \cite{feng2020spatio} (URFD) & 91.4 &-&-&GM&93.1& rt&PP& 94.8  \\ \hline
				
				Feng et al. \cite{feng2020spatio} (MCFD)& 91.6 &93.5 &-  &-  &-  &-  &-  &-  \\ \hline		
				Romaissa et al. (i) \cite{romaissa2020fall} &90 &- &84.60&GM&0.90&RT&PvPr&90.00\\ \hline

				Romaissa et al. (ii) \cite{romaissa2020fall} &89.00 &- &84.60&GM&0.89&RT&PvPr&89.00\\ \hline

				Taufeeque et al.  \cite{taufeeque2021multi} &95.62 &- &98.22&GM&92.56&RT&PvPr&89.76\\ \hline
				
				%
				
			\end{tabular}
			\vspace{-3pt}							
		\end{table}

		\begin{table}[h]
			\vspace{-2pt}
			\centering
			\scriptsize
			\caption{Fall Detection using RNN : Optimization Details}
			\label{T_RNNOptimization Details}
			\vspace{-10pt}
				\begin{tabular}{ |p{2.8cm}|l|l|lHH|l|l|} \hline
					
					\textbf{References}&\textbf{Optimizer} &\textbf{MBS}  &\textbf{Learning Rate} &\textbf{Momentum}&\textbf{Weight Decay}&\textbf{Epochs}&\textbf{Classifier}\\ \hline

					Hasan et al.  \cite{hasan2019robust}&Adam &256  &$5\times10^{-4}$&-&-& 350& Softmax\\ \hline	
					
					Jeong et al. \cite{jeong2019human}&Adam & -  & 0.0001&-& -&500&Softmax\\ \hline

					Feng et al. \cite{feng2020spatio}&Adam &15 & 0.0006&-&-&-&Softmax\\ \hline

					Taufeeque et al.  \cite{taufeeque2021multi}	&- &- & -&-&-&-&Softmax\\ \hline
					

				\end{tabular}
				\vspace{-3pt}								
			\end{table}
			
			Jeong et al. (2019) \cite{jeong2019human} proposed an LSTM based fall detection technique for manufacturing industries. Similar to \cite{hasan2019robust}, here also OpenPose was utilized to get the skeleton data. Two datasets, URFD and SDUFall were exploited. The LSTM with 2 stacked layers and 256 hidden-layers features were used. The raw skeleton data was processed to get the speed of the human centerline coordinate (SHCLC). The raw data with SHCLC was used to get better accuracy results.

			Another work for fall detection using attention-guided LSTM was proposed by Feng et al. (2020) \cite{feng2020spatio}. Here the objective was to detect the falls in a complex background environment where more than one person can be in a frame. The authors created a complex scene fall dataset for their experiments. They also tested their model on the URFD dataset and MCFD dataset for comparison purposes. At first, they used YOLO v3 to detect pedestrians in the scene. Deep-Sort method \cite{wojke2017simple} was used to track the detected object. Features were extracted from the tracked object using VGG-16 which were fed to attention-guided LSTM for detecting falls.

Romaissa et al. (2020) \cite{romaissa2020fall} proposed a LSTM based fall detection using human body geometry. Down-sampling of the video clip was done using optical flow. The angle and the distance between the vector from the centroid of the head and the center of the hip with respect to the horizontal axis were calculated. These values were used to train an SVM and LSTM for fall classification. The Le2i FDD fall dataset was exploited. Two approaches were used: (i) Using angles, distance, Resnet50, SVM, and (ii) Using angles, distance, LSTM. Taufeeque et al. (2021) \cite{taufeeque2021multi} introduced a multi-person and multi-camera-based fall detection technique. They used the UP-Fall dataset which contains data captured by two cameras. Since UP-Fall does not have multi-person data, they also created a dataset using two cameras that contain multi-person. First of all, human pose estimation was done using OpenPifPaf \cite{kreiss2019pifpaf}. Then, multi-person was tracked by mapping keypoints using Gale-Shapley algorithm \cite{gale1962college}. After that, geometrical (angle, aspect ratio, etc.) features were extracted. Then, LSTM was used to detect falls from the extracted features. If a fall was detected then some post-processing was done to reduce false positives.

\subsection{Auto-encoder based Techniques}
In this section fall detection using auto-encoders is reviewed. The details about the experiments discussed in this section are summarized in Table \ref{T_AE_FallDetection}, \ref{T_AE_EvaluationMetrics} and \ref{T_AEOptimization Details}. In the year 2018, Doulamis and Doulamis  \cite{doulamis2018adaptive} proposed a self-adaptable fall detection system using a sparse auto-encoder. This work considered abrupt visual changes like shadows, illumination, background, etc. This model adapts itself to the changing conditions while preserving the current knowledge. At first, humans were detected from the visuals. To do so, sparse auto-encoders with some supervised classification methods were used for the training of the network. To make the network adaptable, approximate methods were used to automatically select the most useful data of the current environment. To conduct the experiment, two datasets, one form mitseva et al.  \cite{mitseva2009isisemd} and one created by the authors themselves, were utilized. The accuracy reported on these two datasets was 85\% and 93\% respectively. 
\begin{table}[h]
	\centering
	\scriptsize
	\vspace{-3pt}
	\caption{Fall Detection using Auto-encoder : Basic Details}
	\label{T_AE_FallDetection}
	
	\vspace{-7pt}	
	\setlength\tabcolsep{1.5pt}	
	\begin{tabular}{|p{3.7cm}|p{1.1cm}|p{2.8cm}|p{3cm}|p{2.4cm}|} \hline
		
		\textbf{Reference}&\textbf{Sensor} &\textbf{Dataset}  &\textbf{Technique}&\textbf{Framework} \\ \hline

		Doulamis et al. (2018) \cite{doulamis2018adaptive} &- & ISESMD \cite{mitseva2009isisemd}, SCDS&Sparse auto-encoder &OpenCV\\ \hline
		
		Nogas et al. (2018) \cite{nogas2018fall}&Thermal  & TSFD \cite{vadivelu2016thermal}& Conv. LSTM AE&- \\ \hline	
		
		Elshwemy et al. (2020) \cite{elshwemy2020new}& Thermal& TSFD \cite{vadivelu2016thermal} & SRAE, ConvLSTM& Keras, Tensorflow\\ \hline

		Nogas et al. (2020) \cite{nogas2020deepfall}  & Depth & TSFD, SDU, URFD&DeepFall, DSTCAE &- \\  \hline

	\end{tabular}
	\vspace{-3pt}	
\end{table}
%
%

%
%
\begin{table}[h]
\scriptsize
\caption{Fall Detection using Auto-encoder : Evaluation metrics}
\label{T_AE_EvaluationMetrics}
\vspace{-5pt}									
	\centering									\begin{tabular}{ |l|HHlHH|HlH|} \hline
		
		\textbf{Reference}&\textbf{Sensitivity} &\textbf{Specificity}  &\textbf{Accuracy}&\textbf{GM}&\textbf{F Score} &\textbf{RT}&\textbf{ROC}&\textbf{Precision}\\ \hline
		
		Doulamis et al. \cite{doulamis2018adaptive} (ISESMD) &-&-&85&-&-&-&-&-  \\\hline
		Doulamis et al. \cite{doulamis2018adaptive} (SCDS) &-&-&93 &-&-&-&-&- \\ \hline		
		Doulamis et al. \cite{doulamis2018adaptive} (Average) &-&-&89 &-&&-&-&- \\ \hline

		Nogas et al.  \cite{nogas2018fall}& - & - & - &- &-&-&0.83 & - \\ \hline				
		Elshwemy et al.  \cite{elshwemy2020new}&-&-&-&gm&-&r-&0.97&- \\ 	\hline	
		%
		
	\end{tabular}
	\vspace{-1pt}			
\end{table}

	%
	%
	
%
\setlength\tabcolsep{1.5pt}											
\begin{table}[h]
	\vspace{-1pt}
	\scriptsize
	\centering
	\caption{Fall Detection using Auto-encoder : Optimization Details}
	\vspace{-7pt}
	\label{T_AEOptimization Details}
		\begin{tabular}{ |p{2.9cm}|p{1.5cm}|p{1.7cm}HHH|p{.9cm}H|p{.9cm}|p{4cm}|} \hline
			
			\textbf{Reference}&\textbf{Optimizer} &\textbf{MBS}  &\textbf{LR}&\textbf{Momen-tum}&\textbf{Weight Decay}&\textbf{Epoch}&\textbf{Classifier}&\textbf{DO }& \textbf{Activation Function}\\  \hline
			
			Nogas et al.  \cite{nogas2018fall}  &Adadelta &16 & LR&Momen &Weight D&50 &clsfr&-&-\\ \hline 		
			
			Elshwemy et al.  \cite{elshwemy2020new}& Adadelta &32&lr&mt&wd&30&cls&-&-\\ \hline

			Nogas et al.  \cite{nogas2020deepfall} &Adadelta &16  & LR&Momen-tum&Weight D& 500 &clsfr& 0.25 & ReLU (Encod.), tanh (Decod.)\\ \hline

		\end{tabular}
		\vspace{-1pt}												
	\end{table}

	Nogas et al. (2018) \cite{nogas2018fall} introduced a thermal image based fall detection method using convolutional LSTM auto-encoder. TSFD dataset was used with data augmentation. The dataset was normalized by dividing each pixel value by 255 to get the value within the range of [0, 1] and subsequently subtracting the resultant pixel value by the mean of the pixels values per frame to get it in the range of [-1, 1]. Elshwemy et al. (2020) \cite{elshwemy2020new} proposed a Spatio-temporal residual auto-encoder (SRAE) model for fall detection using TSFD dataset. A ConvLSTM was utilized for this experiment. Nogas et al. (2020) \cite{nogas2020deepfall} proposed a similar method related to \cite{nogas2018fall}.  Fall detection was considered as anomaly detection from ADL.  Deep spatio-temporal convolutional auto-encoder (DSTCAE) was used to extract spatial and temporal features from ADL. Frames were encoded by the DSTCAE using 3D Conv and 3D Max pooling. Decoding was done using 3D UpSampling and 3D Convolution. Thermal and depth cameras were used which can protect the privacy of the subject.

\subsection{MLP based Techniques}

A significant work for fall detection using MLP was proposed by Safarzadeh et al. (2019) \cite{safarzadeh2019real}. They introduced a two-stage real-time fall detection system. In stage 1, pose estimation was done using a method proposed by Wei et al. \cite{wei2016convolutional}. After pose estimation, an MLP was used to classify these poses as falls or not falls. If a fall is detected an SMS is sent to the caretakers for the necessary action. The prototype system was built using the Arduino Uno board. To detect the joint even in a lying position, for each image extra two images have been generated by rotating it by an angle of -90 and 90 degrees. Two fully connected layer MLP with Relu activation functions was used. The authors created their own dataset for their experiments. The dataset contains 250 images in lying positions in different poses and 250 images in standing, squatting, sitting, and other non-lying positions. Adam optimizer was used for the training. The network was trained till 100 epochs. The maximum accuracy reported is 94\%. Ramirez et al. (2021) \cite{ramirez2021fall} introduced a fall detection method using human skeleton extraction.

\begin{table}[h]
	\centering
	\vspace{-1pt}
	\scriptsize
	\caption{Fall Detection using MLP : Basic Details}
	\label{T_MLP_FallDetection}
	\vspace{-5pt}	
		\begin{tabular}{|p{3.8cm}|l|l|lH|} \hline
			
			\textbf{References}&\textbf{Sensor} &\textbf{Datasets}  &\textbf{Techniques}&\textbf{Framework} \\ \hline

			Safarzadeh et al. (2019) \cite{safarzadeh2019real}&RGB &SCDS &Pose estimation, &- \\ \hline
			
			Ramirez et al. (2021) \cite{ramirez2021fall} &RGB &UP-Fall&Skeleton extraction&- \\  \hline
			
		\end{tabular}
		\vspace{-1pt}		
	\end{table}

	\begin{table}[h]
		\centering
		\vspace{-1pt}
		\scriptsize
		\caption{Fall Detection using MLP : Evaluation metrics}
		\vspace{-7pt}
		\label{T_MLP_EvaluationMetrics}
		
			\begin{tabular}{ |p{2.9cm}|p{1.7cm}|p{1.7cm}|p{1.6cm}H|l|HHl|} \hline
				
				\textbf{References}&\textbf{Sensitivity} &\textbf{Specificity}  &\textbf{Accuracy}&\textbf{GM}&\textbf{F Score} &\textbf{RT}&\textbf{PvPr}&\textbf{Precision}\\ \hline
				
				Safarzadeh et al.  \cite{safarzadeh2019real}&- &- &94& GM&-&Yes&PvPr&-\\ \hline	
				Ramirez et al.  \cite{ramirez2021fall}&94.57$\pm$ 1.15 &98.21 $\pm$ 0.29 &97.39$\pm$ 0.10&GM&94.21$\pm$ 0.27&RT&PvPr&93.87$\pm$ 0.85\\ 	\hline
				
				
			\end{tabular}
			\vspace{-3pt}	
		\end{table}

				%
				%
				%
				%
				%
				%
				%
				%
			%
			%

\subsection{Hybrid model based Techniques}
In many works, more than one type of DL model was used. Those works have been reviewed in this section as hybrid moles. The details about the experiments discussed in this section are summarized in Table \ref{T_HybFallDetection}, \ref{T_Hyb_EvaluationMetrics} and \ref{T_HybridOptimization Details}.

Feng et al. (2014) \cite{feng2014deep} in their work used a single USB camera to capture the images. The background of the image was removed using a codebook background subtraction algorithm to extract the human body image. Some post-processing was applied in this extracted foreground image to remove noises due to shadows etc. Extracted binary silhouettes were fed into deep learning classifiers. Then, the silhouettes were compared with four postures namely standing, sitting, bending, and lying, to detect the falls. For classification, both deep belief network (DBN) and restricted boltzmann machine (RBM) were used. A detection rate of 86\% and a false detection rate of 3.7\% were reported.

\begingroup
\vspace{-8pt}
\scriptsize			
\setlength\tabcolsep{1.5pt}				
\begin{longtable}[h]{|p{3.6cm}|p{1.2cm}|p{2.6cm}|p{4.8cm}|p{2.2cm}|}
	
	\caption{Fall Detection using Hybrid Models : Basic Details}
	\vspace{-10pt}
	\label{T_HybFallDetection}	
	\endfirsthead \hline
	\textbf{Reference}&\textbf{Sensor} &\textbf{Dataset}  &\textbf{Technique}&\textbf{Framework} \\ \hline 
	
	\endhead \hline
	\textbf{Reference}&\textbf{Sensor} &\textbf{Dataset}  &\textbf{Technique}&\textbf{Framework} \\ \hline
	
	Feng et al. (2014) \cite{feng2014deep}&RGB  & SCDS&DBN, RBM &-\\ \hline	
	Lu et al. (2017) \cite{lu2017visual}&- &Sport1-M,  MCFD & 3D CNN, LSTM, spatial visual attention &-\\ \hline
	
	Abobakr et al. (2018) \cite{abobakr2018rgb}&Kinect RGB-D & URFD  \cite{kwolek2014human}  &CNN, LSTM&-\\ \hline

	Ma et al. (2019) \cite{ma2019fall} & RGB, Thermal &Sports-1M, SCDS &3D-CNN and auto-encoder&- \\ \hline
	
	Zhou and Komuro \cite{zhou2019recognizing}&RGB &HQFSD, Le2i FDD&VAE with 3D CNN, Residual blocks&- \\ \hline

	Cai et al. (2020) \cite{cai2020vision}	&RGB &URFD \cite{kwolek2014human}&HCAE& Tensorflow \\ \hline
	
	Manekar et al. \cite{manekar2020activity}& RGB &SCDS& 3D CNN and LSTM &- \\ \hline		
	Chen et al. \cite{chen2020vision}&RGB &URFD \cite{kwolek2014human}, SCDS&R-CNN, bi-directional LSTM&- \\ \hline		
	Pourazad et al. \cite{pourazad2020non}&RGB &Self Created&Inception v3, LSTM&Keras \\ \hline	
	
	Li et al. \cite{li2020multi}&RGB & \cite{auvinet2010multiple, charfi2013optimized, kwolek2014human, baldewijns2016bridging} &Multi-level, DPM&- \\ \hline				
	Mehta et al. (2021) \cite{mehta2021motion}& Thermal &TSFD \cite{vadivelu2016thermal} &Auto-encoder, 3D CNN &- \\  \hline
	Lin et al. \cite{lin2021framework}& - &URFD , Le2i FDD&RNN, LSTM, GRU, OpenPose& Cafe, Keras 2.3 \\  \hline
	Apicella and Snidaro \cite{apicella2021deep}&RGB &URFD, MCFD&Pose estimation, CNN, LSTM&- \\  \hline
	

\end{longtable}
\vspace{-10pt}	
\endgroup

\begin{figure}[h]
	\vspace{-3pt}
	\centering
	\includegraphics[scale=0.65]{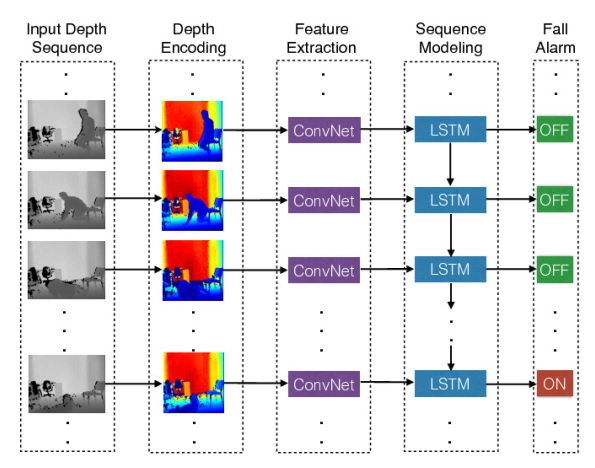}
	\vspace{-10pt}
	\caption{Overview of the  fall detection system  as proposed by Abobakr et al. from \cite{abobakr2018rgb}}
	\label{F_Abobakr18}
	\vspace{-3pt}
\end{figure}

Lu et al. (2017) \cite{lu2017visual} developed a three dimensional convolutional neural network (3D CNN) based fall detection system. Kinetic data were used for the training of the automatic feature extractor. To focus on the key regions, an LSTM based visual attention method was used. The input videos were split into clips. Each clip was of 16 frames with some overlapping frames. The experiment was done with 1, 4, and 8 overlapping frames. The results of the experiment are shown in Table \ref{T_ResultLu}. 3D CNN and LSTM were trained separately. \textbf{Sport-1M} dataset was used to pre-trained the 3D CNN. The 3D feature cube generated through the pre-trained 3D CNN was used as the input to train the LSTM based attention model. 
\begin{table}[h] 
	\vspace{-5pt}
	\scriptsize
	\caption{Results of the experiment reported by Lu et al. \cite{lu2017visual}}
	\vspace{-10pt}
	\label{T_ResultLu}
	\centering
	\begin{tabular}{|c|c|c|c|}\hline
		
		\textbf{No of overlapping frames}&\textbf{Sensitivity} &\textbf{Specificity}  &\textbf{Accuracy}\\ \hline
		
		1& 96.65& 99.85 & 99.73\\ \hline
		4& 86.21&99.56&99.07\\ \hline
		8&65.57&98.37& 97.27\\ \hline
		\textbf{Average}& \textbf{82.81}& \textbf{99.26 }&\textbf{98.69}\\ \hline
		
	\end{tabular}
	\vspace{-3pt}
\end{table}
A similar work to \cite{lu2017visual} was done by Lu et al. (2018) \cite{lu2018deep}. Abobakr et al. (2018) \cite{abobakr2018rgb} proposed a fall detection system using \textbf{depth images} captured by MS Kinect RGB-D sensor \cite{zhang2012microsoft}. This system used deep CNN and RNN to detect falls. Deep CNN was used to extract the visual features from the input frame sequences. Authors followed the residual learning approach (ResNet) \cite{he2016deep} for the  CNN. On the top of the ResNet LSTM \cite{hochreiter1997long} recurrent neural network was used. LSTM gave the temporal dynamics information which discriminates an event as ``fall" or ``no fall". URFD was exploited for the training and evaluation of this system. 0.0005 was used as initial LR with a decaying factor of 10 after every 30 epochs. The overview of the proposed fall detection system is shown in Figure \ref{F_Abobakr18}.

Ma et al. (2019) \cite{ma2019fall} proposed a privacy-preserving fall detection system using 3D CNN (C3D) and auto-encoder (AE). For preserving privacy they used an optical level anonymous image sensing system (OLAISS) \cite{zhang2014anonymous}. The advantage of using OLAISS is that it hides (blackouts) the facial regions at the time of capturing instead of doing this in post-processing. C3D was pre-trained using the Sports-1M dataset. To evaluate their performance they created their own dataset with two cameras one for side view and another one for front view recording simultaneously. Zhou and Komuro (2019) \cite{zhou2019recognizing} proposed a fall detection system using a variational auto-encoder (VAE) \cite{kingma2013auto} with 3D CNN residual blocks \cite{hara2017learning}. Here, the region with a human were extracted (cropped) and aligned at the left shoulder. They used unsupervised techniques with some weakly labeled data. Reconstruction error (mean square error) was used to detect falls. If the error was within the normal range, it was detected as normal ADL. If it was outside of the range, it was a fall. AlphaPose \cite{fang2017rmpe} was used to extract the region of human motion. The HQFSD and the Le2i FDD were used. 8 residual blocks of ResVAE-18 or 16 residual blocks of ResVAE-34 were used in the encoder and decoder of the VAE. For training the ResVAE-18, and ResVAE-34, the 12266 samples (ADL) of the HQFSD dataset were exploited. The testing of the ResVAE-18 and ResVAE-34 was done using 282 fall samples and 300 ADL samples from the HQFSD dataset and 130 fall samples and 200 ADL samples from the Le2i dataset. 

\begin{table}
	\centering
	\vspace{-10pt}
	\scriptsize
	\caption{Fall Detection using Hybrid models : Evaluation Metrics}
	\label{T_Hyb_EvaluationMetrics}
	\vspace{-10pt}
		\begin{tabular}{ |p{6cm}|l|l|lH|lHH|l|} 
			\hline
			\textbf{References}&\textbf{Sensitivity} &\textbf{Specificity}  &\textbf{Accuracy}&\textbf{GM}&\textbf{F Score} &\textbf{RT}&\textbf{PvPr}&\textbf{Precision}\\ \hline

			Feng et al. \cite{feng2014deep}&-&-&86& 15&&No&-&-\\ \hline		
			
			Lu et al. \cite{lu2017visual}& 82.81& 99.26 & 98.69 &RT&-&PvPr&-&- \\ \hline
			Abobakr et al. \cite{abobakr2018rgb}&100 &97&98&GM&-&Yes&Yes&-\\ \hline

			Ma et al.  \cite{ma2019fall}&0.933 &0.928 &-&GM&-&RT&Yes&-\\ \hline

			Zhou and Komuro (ResVAE-18-HQFSD) \cite{zhou2019recognizing}&- &- & 88.7& GM& - &RT&PvPr&-\\ \hline
			Zhou and Komuro (ResVAE-18-Le2i) \cite{zhou2019recognizing}&- &- & 87.3&GM&-&RT&PvPr&-\\ \hline
			Zhou and Komuro (ResVAE-34-HQFSD) \cite{zhou2019recognizing}&- &- & 82&GM&-&RT&PvPr&-\\ \hline
			Zhou and Komuro (ResVAE-34-Le2i) \cite{zhou2019recognizing}&- &- & 87.6&GM&- &RT&PvPr&-\\ \hline
			Cai et al. \cite{cai2020vision}& 100& 93& 96.2&GM&96&rt&pp& 92.3\\ \hline	
			
			Manekar et al. (3D CNN) \cite{manekar2020activity}&-& - & 95.3&&-&&&-\\ \hline
			Manekar et al. (CNN-LSTM) \cite{manekar2020activity}&-& - & 91.63&&-&&&-\\ \hline			
			Chen et al. (URFD) \cite{chen2020vision}& 91.8 & 100 & 96.7&GM& 94.8&RT&PvPr& 100\\ \hline	
			Chen et al. (Self dataset) \cite{chen2020vision}& 92.3 &- &-&GM& 94.8&RT&PvPr& 98.1\\ \hline	
			Pourazad et al. \cite{pourazad2020non}&- &- & 87&GM&-&RT&PvPr&-\\ \hline

			Apicella and Snidaro  \cite{apicella2021deep}	&80.1 &SP &81.8&GM&80.5&RT&PvPr&80.9\\ \hline 
			
		\end{tabular}
		\vspace{-3pt}	
	\end{table}

	\begin{table}[h]
		
		\scriptsize
		\caption{Fall Detection using Hybrid Models : Optimization Details}
		\vspace{-10pt}
		\label{T_HybridOptimization Details}
			\setlength\tabcolsep{1.5pt}
			\begin{tabular}{ |l|p{1.3cm}|p{1cm}|p{1.4cm}|p{1cm}|p{1cm}|l|p{1.4cm}|l|l|} 
				\hline 
				\textbf{Reference}&\textbf{Optimiz.} &\textbf{MBS}  &\textbf{L.R.}& \textbf{Momt.}&\textbf{W.D.}&\textbf{Epoch}&\textbf{Classifier}&\textbf{DO}&\textbf{AF}\\ \hline

				Lu et al. \cite{lu2017visual}& Adam  & -& -&- &- &- &- &- &- \\ \hline
				Abobakr et al. \cite{abobakr2018rgb}&SGD &160 frames  &0.0005& 0.9& 0.0001& 100 & Softmax  &- &-\\ \hline
				
				Ma et al.  \cite{ma2019fall}& AdaGrad &28  & 10\^-4&- & -&10&Softmax and SVM  &- &-\\ \hline

				Zhou and Komuro (ResVAE-18) \cite{zhou2019recognizing}&Adam &32 & -&-&-&500&-&-  &-\\ \hline
				Zhou and Komuro (ResVAE-34) \cite{zhou2019recognizing}&Adam &24 & -&-&-&500&-&- &-\\ \hline
				Manekar et al. (3D CNN) \cite{manekar2020activity} &Adam&128&.0001&&-&300&Softmax&0.25 &- \\ \hline
				Manekar et al. (CNN-LSTM) \cite{manekar2020activity}& Adam&& 0.00001&-&-& 60&-&- &-\\ \hline		
				Mehta et al. \cite{mehta2021motion}&SGD, Adadelta &- & 0.0002&-&-&300&-&- &-\\ \hline 
				Lin et al.  \cite{lin2021framework}&Adam & 916& 0.1, 0.01, 0.001&-&-&500&-&-& Tanh\\ \hline

				Apicella and Snidaro  \cite{apicella2021deep}&Adam &- & 0.0001&-&-&1000&Sigmoid&0.2& ReLU\\ \hline
				
				
			\end{tabular}
			
		\end{table}

		Cai et al. (2020) \cite{cai2020vision} proposed a multi-task hourglass convolutional auto-encoder (HCAE) based fall detection system. Multi-scale features were extracted using an hourglass residual unit (HRU). Manekar et al. (2020) \cite{manekar2020activity} proposed a fall detection method based on 3D CNN and LSTM. They used two techniques, one using 3D CNN and another one 3D CNN combined with LSTM. They created a 360-degree dataset using an omnidirectional camera. Chen et al. (2020) \cite{chen2020vision} proposed a attention guided bi-directional LSTM based fall detection method in complex scenes environment. URFD and a self-created dataset were exploited.  To make the background complex Gaussian noise was added into the authors' created dataset. For background subtraction, Mask R-CNN \cite{8372616} was used. After removing the background, features were extracted using the output of the last Conv layer of VGG-16. Finally, Extracted features were fed into attention-guided Bi-directional LSTM for fall detection.
		
		Pourazad et al. (2020) \cite{pourazad2020non} introduced a fall detection system using CNN and LSTM. A self-created RGB video dataset was used. A pre-trained  Inception V3 (modified) network on the ImageNet dataset was used to extract features. The last output layer of the inception model was replaced by a 2048 unit FC layer. The extracted features were fed to three-layer LSTM for final fall detection. Kinects camera was used for video capturing. Li et al. (2020) \cite{li2020multi} classified a fall detection technique into three types \cite{li2020multi}. Frame-level fall detection (FLFD), Sequence-level fall detection (SLFD) and video-level fall detection (VLFD). Li et al. \cite{li2020multi} presented a multi-level dynamic pose motion (DPM) based fall detection method using CNN-LSTM. They labeled the five public datasets frame-wise.

		Mehta et al. (2021) \cite{mehta2021motion} proposed a thermal-image-based fall detection system using two-channel 3D auto-encoders and 3D CNN. First of all, region of interest (ROI) extraction and optical flow computation were done from the thermal input frames. Thermal frames and optical flow frames with ROI masking were fed to thermal auto-encoder and flow auto-encoder respectively. The outputs of these two auto-encoders were given to the 3D-CNN thermal and flow discriminator. Finally, the fall was detected using the Joint discriminator of thermal and flow discriminator. Region-based fully convolutional network (R-FCN) \cite{dai2016r}, Otsu thresholding \cite{otsu1979threshold}, Kalman filtering were used for person detection, Contour box localization, and tracking respectively.

		Lin et al. (2021) \cite{lin2021framework} presented OpenPose based fall detection method using RNN, LSTM and gated recurrent unit (GRU). At first, sequences of continuous images were extracted from the input dataset. Then, the skeletons were generated from these sequences of images. After that, some pre-processing was done. Then, these pre-processed data were trained and tested. The URFD and Le2i FDD datasets were exploited after mixing these two datasets. If some part was missing due to overlapped or obscured, the linear interpolation was tried. The performance of linear interpolation was not so good. The results are shown in Table \ref{T_ResultLin}.
		
		\begin{table}[h] 
			\vspace{-10pt}
			\scriptsize
			\caption{Results of the experiment reported by Lin et al. \cite{lin2021framework}}
			\label{T_ResultLin}
			\vspace{-10pt}
			\centering
			\begin{tabular}{|l|l|l|l|} \hline
				
				\textbf{Metric}&\textbf{RNN} &\textbf{LSTM}  &\textbf{GRU}\\ \hline
				
				Sensitivity (RP-Normalization)& 93.7& 100&96.4\\ \hline
				Sensitivity (Linear Interpolation + RP-Normalization)& 92.8&95.5&96.4\\ \hline
				Specificity (RP-Normalization)& 84.8& 96.4&98.2\\ \hline
				Specificity (Linear Interpolation + RP-Normalization)& 87.5 &94.6&98.2\\ \hline 
				Accuracy (RP-Normalization)& 89.2&98.2&97.3\\ \hline
				Accuracy (Linear Interpolation + RP-Normalization)&90.1& 95&97.3	\\ \hline
				
			\end{tabular}
			\vspace{-10pt}
		\end{table}

		Romaissa et al. (2021) \cite{romaissa2021vision} proposed human body geometry-based fall detection method using LSTM and SVM classifier. Two geometric features were used, one was the angle between the line from the center of the head to the center of hip and horizontal axis, and another feature was the distance from the vector forming from the center of the head to the center of the heap and the vector forming from the horizontal axis. URFD and Le2i FDD dataset were exploited. At first, the down-sampling of the input video was done by calculating its optical flow. After that,  manual annotations of the body and head were done. Then, angle and distance were calculated. Finally, falls were detected using a  bidirectional LSTM and an SVM. Different combinations of architectures (Resnet50,  AlexNet) and features (angle only, angle, and distance) were used. The performance using feature images + ResNet50 + SVM is shown in Table \ref{T_Result_Romaissa}.
		
		\begin{table}[h]
			\vspace{-8pt}	
			\scriptsize
			\centering
			
			\caption{Results as reported by Romaissa et al. \cite{romaissa2021vision} }
			\label{T_Result_Romaissa}
			\vspace{-10pt}
			
			\begin{tabular}{|p{1.4cm}|p{1cm}ccc|cccc|}
				
				\hline
				
				\multirow{2}{*}{\textbf{Features} }&\multicolumn{4}{c|}{\textbf{URFD}} &\multicolumn{4}{c|}{\textbf{Le2i}} \\

				& \textbf{Accur.} & \textbf{Precis.}& \textbf{Sens.} & \textbf{F Score}& \textbf{Accur.} & \textbf{Precis.} &\textbf{Sens.} & \textbf{F Score}\\ \hline
				
				Angle	& 0.960 & 1.000& 0.8333 & 0.907& 0.962 & 0.952 & 1.000 & 0.975\\ \hline
				Angle + Distance & 0.960 & 1.000 & 0.900 & 0.947 & 0.962 & 1.000 & 0.950 & 0.974 \\ \hline 
				\textbf{Average} & 0.960 & 1.000& 0.86665& 1.854 &0.962 &0.976 &0.975 & 0.9745\\ \hline

			\end{tabular}
			\vspace{-10pt}	
		\end{table}

		Apicella and Snidaro (2021) \cite{apicella2021deep} presented a fall detection method using LSTM and CNN. At first, pose estimation was done using PoseNet \cite{papandreou2018personlab}. If pose estimation was done accurately then extracted series of poses (20 poses) were fed to LSTM. If pose was not generated successfully or some keypoints (among 17 keypoints) were missing then an additional CNN was used to generate the series of poses. Finally, LSTM classified an activity as fall or no fall.

		%
		%
		%
		
		%
		
		%
		%

\section{Discussions on Limitations and Future Scope}
\label{S_Discussion}
The maximum work for fall detection using DL have been done using CNN followed by hybrid, LSTM, Auto-encoder and MLP as shown in Figure \ref{F_PaperTypes}.	\begin{figure} [h]
	\scriptsize
	\centering
	\begin{tikzpicture}[rotate=0, level 1/.style={sibling distance=25mm,level distance=17mm}, level 2/.style={sibling distance=15mm,level distance=20mm}, level 3/.style={sibling distance=11mm,level distance=40mm},
		every node/.style = {shape=rectangle, rounded corners,
			draw, align=center,
			top color=white, bottom color=blue!20, auto}]]		
		\node [align=center] {\textbf{Classification of papers based on the CNN modle used}}
		child { 
			node [left=-7mm, text width=5 cm]{\textbf{\underline{CNN}}\\ \cite{doulamis2016vision, fan2017deep, nunez2017vision, li2017fall, solbach2017vision, hsieh2017development, iuga2018fall,zhang2018fall, shen2018fall, kong2019robust, el2019reduce, cameiro2019multi, leite2019fall,  brieva2019intelligent, cai2019novel, espinosa2019vision, cai2019fall, kasturi2019human, wu2019skeleton, zheng2019fall,   espinosa2020application, carlier2020fall, yao2020novel, menacho2020fall,  chen2020edge,   ijjina2020human, hader2020automatic,   dichwalkar2020activity, asif2020sshfd, lezzar2020camera, zhang2020human, zhong2020multi,  asif2020privacy, euprazia2020novel, liutwo, chen2020fall, kareem2020using,       abdo2020fall,  chhetri2021deep, chen2021video, killian2021fall, leite2021three, zou2021movement, vishnu2021human, berlin2021vision,cai2021vision, li2021fall, keskes2021vision}} }
		child { 
			node  [left=-7mm, text width=2cm] {\textbf{\underline{LSTM}}\\ \cite{hasan2019robust, jeong2019human, feng2020spatio, romaissa2020fall, taufeeque2021multi}} }
		child { 
			node  [left=-7mm, text width=2cm] {\textbf{\underline{Auto-encoder}}\\ \cite{doulamis2018adaptive, nogas2018fall, elshwemy2020new, nogas2020deepfall}} }
		child { 
			node [left=1mm, text width=1.3cm] {\textbf{\underline{MLP}} \\ \cite{safarzadeh2019real,  ramirez2021fall}} }
		child { 
			node [left=3mm, text width=1.8cm] {\textbf{\underline{Hybrid}} \\ \cite{feng2014deep, lu2017visual, lu2018deep, abobakr2018rgb, ma2019fall, zhou2019recognizing, cai2020vision, manekar2020activity, chen2020vision, pourazad2020non, li2020multi, mehta2021motion, lin2021framework, romaissa2021vision, apicella2021deep} }};	
	\end{tikzpicture}
	\vspace{-7pt}
	\caption{Classification of papers based on the CNN model used}
	\vspace{-10pt}
	\label{F_PaperTypes} 
\end{figure}
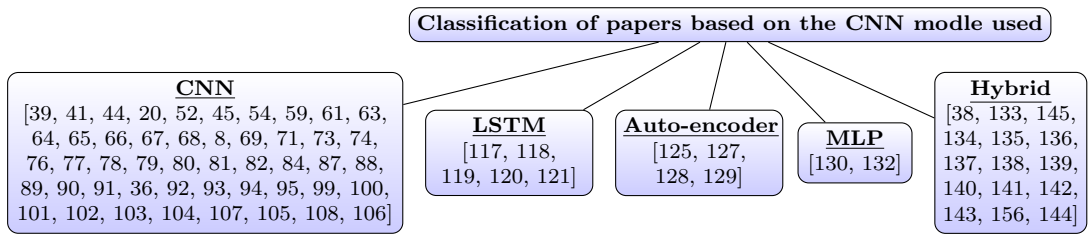								These DL based techniques gives very good results, but some flaw is there. DL based techniques are in genral very data hungry and needs high computation especially for training of the model. It's very difficult to use these models in embedded IoT devices. In IoT devices not all computations is done in these devices locally, lots of files transfer from these devices (edges) to cloud servers and vice versa. This may be issue of security and privacy. The main focus of the works of the reviewed papers is the good results in terms of the metrics mentioned in section \ref{S_EvaMetrics}. Privacy and security should also be the focus of works. Also, majority of the works have been done using RGB inputs which is also a concern of the privacy of the subjects. To protect the privacy of the subjects more wokrs can be done using thermal and infrared sensors. Depth cameras use infrared to capture the subjects. These types of cameras can work in low light also. Here, there is no concern for privacy. Some works were done using multiple cameras (\cite{espinosa2019vision, yao2020novel, taufeeque2021multi} etc.). The advantage of using multiple cameras is that even if one camera is not working or the subject is not in the view of both cameras then also the system will work with the available camera. Multiple cameras system is generally more complex and all cameras must work in a synchronize manner with each other.

\section{Conclusion}
\label{S_Conclusion}
In this paper, we have reviewed the recent (since 2014) developments in the DL based non-intrusive (vision-based) HFD methods. We have described the different metrics which are used to evaluate the performance of these fall detection methods. Sensitivity (recall), specificity, accuracy, F score, precision, and geometric mean were defined with their respective equations. The HFDSs which are publicly available, are also surveyed briefly. A brief description of MCFD, Le2i FDD, URFD, SDUFall, HQFSD, TSFD, SisFall, and UP-Fall datasets have been provided. To review the DL based fall detection methods, we have classified all methods according to the DL model used. We have classified it to CNN, Auto-Encoder, LSTM, MLP, and hybrid. Lots of work have been done in this field which gives good results. Some more work is needed considering the multi-person in a frame and occlusion. More work should be done considering privacy and security also. Thermal and infrared cameras can be utilized for privacy. Edge computing can be useful for providing security.  The datasets which are available have been created by simulating the fall activities by some actors. These datasets do not contain the activities of the actual subjects (patients/ old people). A dataset using thermal or infrared sensors can be created which will contain the activities of the actual subjects. Thermal and infrared sensors will also preserve the privacy of the subjects.

\bibliographystyle{elsarticle-num}
\scriptsize{\bibliography{VHFDSDL_MDPI_bib}}

\end{document}